\documentclass{article}

\PassOptionsToPackage{numbers}{natbib}
\PassOptionsToPackage{sort&compress}{natbib}
\usepackage[final]{neurips}
\usepackage[utf8]{inputenc}
\usepackage[T1]{fontenc}
\usepackage[colorlinks=true, linkcolor=cyan, citecolor=cyan, urlcolor=cyan]{hyperref}
\usepackage{url}
\usepackage{booktabs}
\usepackage{amsfonts}
\usepackage[dvipsnames]{xcolor}
\usepackage{nicefrac}
\usepackage{microtype}
\usepackage{wrapfig}
\usepackage{ulem}
\usepackage{comment}
\usepackage{graphicx}
\usepackage{amsmath}
\usepackage{titletoc}
\usepackage{caption}
\usepackage{array}
\usepackage{xurl}
\usepackage{tabularx}
\usepackage{enumitem}
\usepackage{float}
\usepackage{multirow}
\newcolumntype{L}[1]{>{\raggedright\arraybackslash}p{#1}}

\newcommand{\pokeagent}{\text{PokéAgent }}

\renewcommand{\thefootnote}{$\diamond$}

\title{The PokeAgent Challenge: \\ Competitive and Long-Context Learning at Scale}

\author{
  Seth Karten$^{1,*}$\thanks{Contribution statement in Appendix~\ref{sec:authors}. $^{*}$Equal contribution. Correspondence: \texttt{sethkarten@princeton.edu, grigsby@cs.utexas.edu}} \quad
  Jake Grigsby$^{2,*}$ \quad
  Tersoo Upaa Jr$^{1}$
  \AND
  \parbox{0.95\textwidth}{\centering
  Junik Bae$^{6}$ \quad
  Seonghun Hong$^{14}$ \quad
  Hyunyoung Jeong$^{14}$ \quad
  Jaeyoon Jung$^{14}$ \quad
  Kun Kerdthaisong$^{15}$ \quad
  Gyungbo Kim$^{9}$ \quad
  Hyeokgi Kim$^{14}$ \quad
  Yujin Kim$^{9}$ \quad
  Eunju Kwon$^{9}$ \quad
  Dongyu Liu$^{7}$ \quad
  Patrick Mariglia$^{8}$ \quad
  Sangyeon Park$^{9}$ \quad
  Benedikt Schink$^{12}$ \quad
  Xianwei Shi$^{7}$ \quad
  Anthony Sistilli$^{11}$ \quad
  Joseph Twin$^{13}$ \quad
  Arian Urdu$^{12}$ \quad
  Matin Urdu$^{12}$ \quad
  Qiao Wang$^{10}$ \quad
  Ling Wu$^{10}$ \quad
  Wenli Zhang$^{7}$ \quad
  Kunsheng Zhou$^{7}$ \\[4pt]
  Stephanie Milani$^{3,4}$ \enspace
  Kiran Vodrahalli$^{5}$ \enspace
  Amy Zhang$^{2}$ \enspace
  Fei Fang$^{3}$ \enspace
  Yuke Zhu$^{2}$ \enspace
  Chi Jin$^{1}$}
  \AND
  \parbox{0.95\textwidth}{\centering\mdseries\small
  $^{1}$Princeton \quad $^{2}$UT-Austin \quad $^{3}$CMU \quad $^{4}$NYU \quad $^{5}$Google DeepMind \\
  $^{6}$Team Heatz \quad $^{7}$Team PA-Agent \quad $^{8}$Team FoulPlay \quad $^{9}$Team 4thLesson \quad $^{10}$Team Q \\
  $^{11}$Team Anthonys \quad $^{12}$Team Hamburg \quad $^{13}$Team Porygon2AI \quad $^{14}$Team Deepest \quad $^{15}$Team August 
  }
}

\begin{document}

\maketitle
\renewcommand{\thefootnote}{\arabic{footnote}}

\begin{abstract}
    We present the \pokeagent Challenge, a large-scale benchmark for decision-making research built on Pokémon's multi-agent battle system and expansive role-playing game (RPG) environment. Partial observability, game-theoretic reasoning, and long-horizon planning remain open problems for frontier AI, yet few benchmarks stress all three simultaneously under realistic conditions. \pokeagent targets these limitations at scale through two complementary tracks: our \textit{Battling Track}, which calls for strategic reasoning and generalization under partial observability in competitive Pokémon battles, and our \textit{Speedrunning Track}, which requires long-horizon planning and sequential decision-making in the Pokémon RPG. Our Battling Track supplies a dataset of 20M+ battle trajectories alongside a suite of heuristic, RL, and LLM-based baselines capable of high-level competitive play. Our Speedrunning Track provides the first standardized evaluation framework for RPG speedrunning, including an open-source multi-agent orchestration system that enables modular, reproducible comparisons of harness-based LLM approaches. Our NeurIPS 2025 competition validates both the quality of our resources and the research community's interest in Pokémon, with more than 100 teams competing across both tracks and winning solutions detailed in our paper. Participant submissions and our own baselines reveal considerable gaps between generalist (LLM), specialist (RL), and elite human performance. Analysis against the BenchPress evaluation matrix shows that Pokémon battling is nearly orthogonal to standard LLM benchmarks, measuring capabilities not captured by existing evaluation suites and positioning Pokémon as an unsolved benchmark that can drive RL and LLM research forward. We transition from the NeurIPS 2025 competition to a living benchmark by releasing a live leaderboard for Battling and self-contained evaluation for Speedrunning at \url{https://pokeagentchallenge.com}.
    \end{abstract}

\section{Introduction}
\label{sec:intro}

Partial observability, game-theoretic reasoning, and long-horizon planning are core challenges in sequential decision-making, yet few existing benchmarks stress all three simultaneously under realistic conditions. Standard testbeds tend to isolate one axis---imperfect-information games emphasize equilibrium computation in short episodes, while open-ended environments test exploration but lack adversarial opponents. Pokémon is an environment that combines all three: competitive battles require reasoning under hidden information against a strategic adversary, while single-player campaigns demand thousands of cumulative decisions spanning exploration, resource management, and combat over extended horizons. With approximately $10^{564}$ possible battle states (see Appendix~\ref{sec:state-space-derivation}), team building across 1,000+ species with diverse movesets and abilities, and a competitive metagame that evolves continuously, Pokémon is more complex and dynamic than most existing benchmarks.

In 2025, Pokémon gained significant interest for evaluating frontier AI systems. Claude Plays Pokémon ~\citep{anthropic2025visible} demonstrated extended thinking capabilities over 35,000 actions to complete a small section of the game, Gemini 2.5 Pro completed the entire game of Pokémon Blue in 406 hours~\citep{gemini2p5report,zhang2025geminiplayspokemon}, and OpenAI's GPT-5 finished the game in 6,470 steps~\citep{engadget2025gptplayspokemon}.
These demonstrations reinforced Pokémon's suitability as an AI testbed, but the efforts were fragmented---different games (Red, Blue, Crystal, Emerald), different harnesses, and different evaluation criteria made meaningful comparison impossible. Was Gemini 3 Pro's 173-hour completion better than Claude Opus 4.6 reaching Victory Road? Did GPT-5's step count account for the same mechanics? By conflating harness with model capability, it became impossible to attribute success to the agent architecture, the underlying model, or hardcoded assumptions that simplified perception. The importance of standardized evaluation in games AI is well established: the Arcade Learning Environment~\citep{bellemare2013arcade} catalyzed a decade of RL progress~\citep{mnih2015human}, while MineRL~\citep{guss2019minerl, milani2023solvingfuzzytaskshuman, shah2022retrospective} demonstrated shared benchmarks for open-ended environments. Pokémon demands---and now receives---similar standardization.

Pokémon also offers a distinctive form of out-of-distribution evaluation. While extensive Pokémon knowledge exists in pretraining corpora---move types, damage formulas, competitive tier lists---translating that latent knowledge into effective multi-turn sequential decision-making under partial observability is fundamentally different from the recognition and retrieval tasks where data contamination typically inflates performance. Moreover, the competitive metagame shifts continuously as the player community develops new strategies and the game's governing body rebalances tiers---creating natural distribution shifts that test an agent's ability to adapt rather than memorize.

We present the \pokeagent Challenge, a standardized evaluation framework for Pokémon-playing AI agents. The benchmark features two complementary tracks: \textbf{Competitive Battling} evaluates strategic reasoning under partial observability in two-player competitive Pokémon, while the \textbf{RPG Speedrunning} tests long-horizon planning in completing Pokémon Emerald as quickly as possible.

The NeurIPS 2025 \pokeagent Challenge confirmed the benchmark's difficulty and drew strong community engagement: 100+ teams submitted solutions across both tracks and 650+ researchers joined the competition Discord for technical exchange. The competition produced novel methods---including Scripted Policy Distillation for RPG play and iterative offline RL with dynamic data weighting---and served as the first large-scale competitive testbed for approaches such as root-parallelized MCTS in imperfect-information battling~\citep{FoulPlay}, while revealing a capability hierarchy: specialist RL and search methods dominated LLM approaches in Competitive Battling, and no raw frontier model achieved non-trivial progress in Speedrunning without a sophisticated harness. These gaps remain far from closed.

Our contributions include: (1)~the \pokeagent Challenge, a two-track evaluation framework pairing competitive battling (via Pokémon Showdown) with RPG speedrunning (via Pokémon Emerald), with standardized infrastructure for fair comparison across RL, LLM, and hybrid approaches; (2) the largest publicly available Pokémon battle dataset---comprising 4M human demonstrations and 18M synthetic battles, plus 200K+ curated competitive teams; (3)~baselines spanning heuristic bots, RL agents, and harness LLM agents, alongside the first open-source multi-agent orchestration system for long-horizon RPG play; (4)~empirical validation through the NeurIPS 2025 \pokeagent Challenge (100+ competitors, 100K+ battles, top methods in Appendix~\ref{sec:participant-methodologies}), with results revealing substantial gaps between generalist LLM, specialist RL, and elite human performance, and orthogonality analysis showing that Pokémon battling captures capabilities not predicted by the 49 benchmarks in the BenchPress evaluation matrix~\citep{papailiopoulos2026benchpress}; and (5)~a living benchmark with a live Battling and Speedrunning leaderboard and self-contained Track~2 evaluation at \url{https://pokeagentchallenge.com}.

\section{Related Work}

\begin{wrapfigure}{r}{0.48\textwidth}
\centering
\vspace{-4mm}
\includegraphics[width=1.05\linewidth]{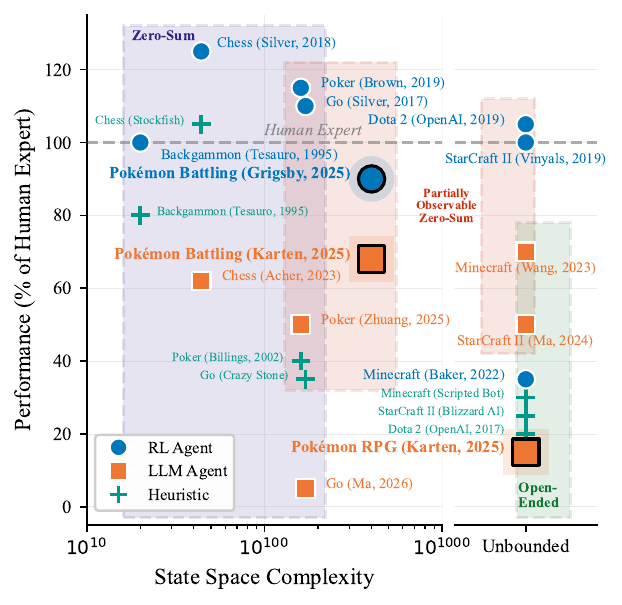}
\vspace{-6mm}
\caption{\textbf{Game Benchmarks.} Pokémon creates vast partially observable state spaces (see Appendix~\ref{sec:state-space-derivation}). Data from \cite{tesauro1995temporal, silver2018general, silver2017mastering, brown2018superhuman, brown2019superhuman, billings2002challenge, vinyals2019grandmaster, openai2019dota, baker2022video, wangvoyager, acher2024gpt4chess, ma2026mixing, yan2025pokerbench, ma2024llmstarcraft}.}
\label{fig:benchmarks}
\vspace{-6mm}
\end{wrapfigure}

\paragraph{Game AI Benchmarks.}
Traditional benchmarks rapidly saturate, but adversarial games resist this by forcing continuous adaptation. Game AI has driven major advances: superhuman board games~\citep{silver2018general}, imperfect-information poker~\citep{brown2018superhuman, brown2019superhuman}, grandmaster-level StarCraft~II~\citep{vinyals2019grandmaster}, and human-level Diplomacy combining language models with strategic reasoning~\citep{meta2022cicero}. As Figure~\ref{fig:benchmarks} shows, RL achieves superhuman performance in fully observable settings, but this margin erodes in stochastic, partially observable environments~\citep{samuel1959}, and LLM agents consistently lag specialist RL and search systems~\citep{kaggle2025gamearena}. Hanabi~\citep{bard2020hanabi} and FightLadder~\citep{li2024fightladder} advance imperfect-information and competitive evaluation; NetHack~\citep{kuttler2020nethack} and BALROG~\citep{paglieri2024balrog} benchmark long-horizon reasoning for RL and LLM agents. However, none combines adversarial reasoning with long-horizon planning at scale, and most rely on symbolic state representations rather than visual perception. Pokémon offers a unique combination: an enormous partially observed state space, a visual RPG requiring pixel-level perception, and a massive active player base generating continuous human data and an evolving competitive metagame. The gaps between AI and humans, and between specialist and generalist AI, remain wide open.

\paragraph{NeurIPS Game AI Competitions.}
Standardized competitions have been important for advancing game AI. Neural MMO~\citep{liu2023neurips, suarez2023neural} benchmarks multi-agent cooperation, Lux AI~\citep{tao2024lux} targets resource management with shifting dynamics, MineRL~\citep{guss2019minerl, milani2023solvingfuzzytaskshuman, shah2022retrospective} addresses long-horizon planning in open worlds. While each advances a specific research axis, none combines adversarial partial observability with long-horizon planning at the scale of a living competitive ecosystem. \pokeagent bridges this gap: its dual-track design jointly evaluates RL and LLM approaches in high-stakes competitive play (battling) and extended sequential decision-making over thousands of steps (speedrunning), providing complementary stress tests that no single existing benchmark covers.

\paragraph{Prior Work on Pokémon AI.}
Our prior work introduced PokéChamp~\citep{karten2025pok}, combining minimax search with LLMs, and Metamon~\citep{grigsby2025human}, training RL agents on millions of human and self-play battles. On the RPG side, Puffer~\citep{pleines2025pokemon} demonstrated RL-based completion of Pokémon Red, while demonstrations from Claude~\citep{engadget2025claudeplayspokemon,anthropic2025visible}, Gemini~\citep{gemini2p5report,zhang2025geminiplayspokemon, zhang2025gemini3playspokemon}, and GPT~\citep{engadget2025gptplayspokemon} showed both the strengths and limitations of frontier models. However, each effort produced individual systems rather than reusable benchmark infrastructure: none established standardized evaluation, public leaderboards, or multi-track design for fair cross-paradigm comparison. The \pokeagent Challenge extends these efforts into a unified evaluation framework.

\section{Competitive Battling Track}

\subsection{Battling Environment Design}
\begin{wrapfigure}{r}{0.30\linewidth}
    \centering
    \vspace{-10mm}
    \includegraphics[width=.85\linewidth]{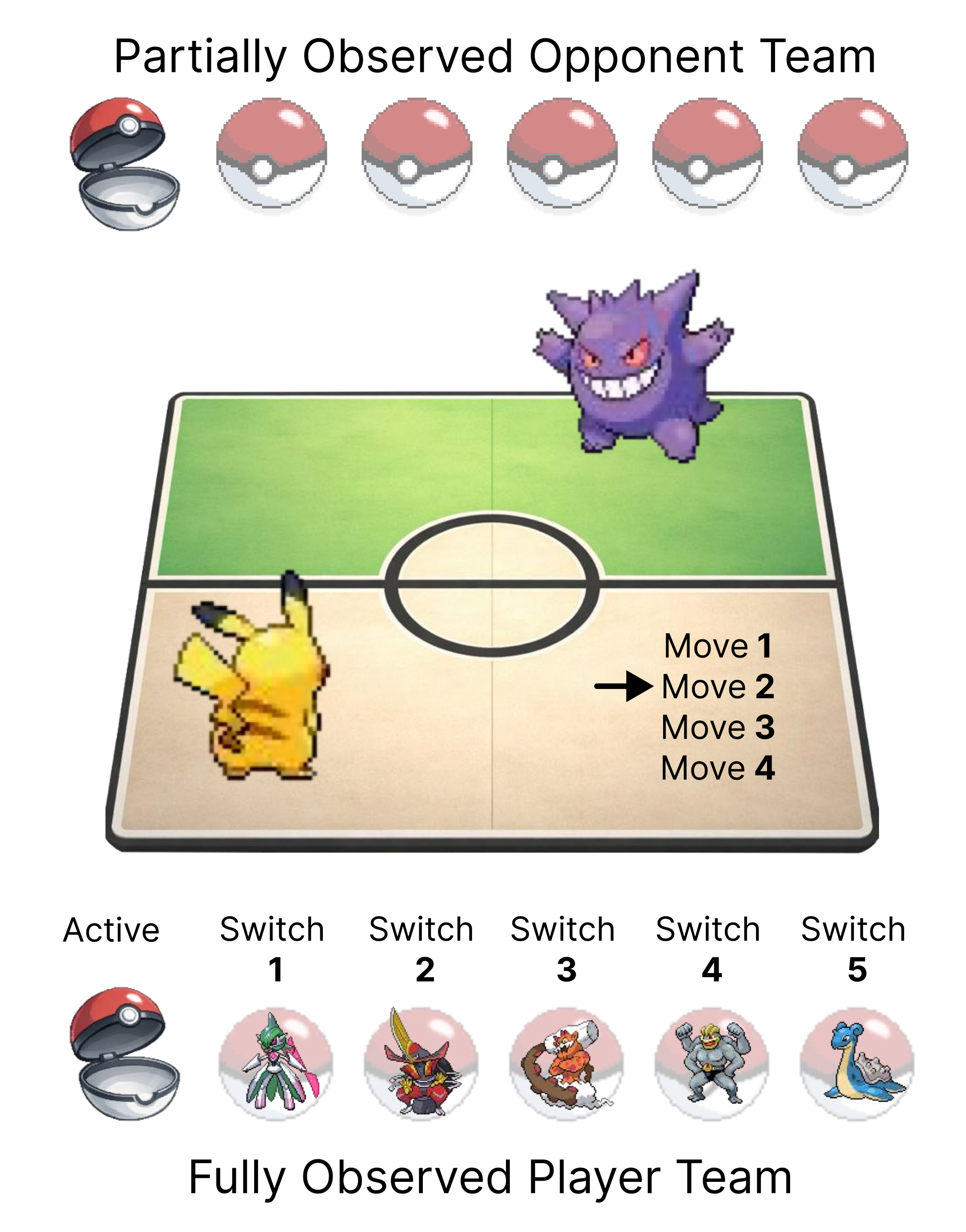}
    \vspace{-2mm}
    \caption{\textbf{Pokémon Battling.}}
    \vspace{-8mm}
    \label{fig:battle}
\end{wrapfigure}

\href{https://play.pokemonshowdown.com}{Pokémon Showdown} is an open-source simulator that transforms Pokémon's turn-based battle mechanics into a standalone competitive game with thousands of daily players. Formally, battles are two-player, zero-sum, stochastic games with imperfect information and simultaneous action selection. Their imperfect information primarily stems from team construction: each player drafts a team from a vast design space, and key aspects of the opponent’s team remain hidden until revealed through play. On each turn, players select from \textasciitilde9 actions (Figure \ref{fig:battle}), with battles lasting \textasciitilde20--100 turns. Action outcomes are stochastic and can lead to a long tail of rare events that abruptly swing state values. The combination of randomness, hidden information, team diversity, long-horizon planning, and evolving rules presents a significant challenge, and evaluating progress is difficult: existing work typically relies on disjoint baselines or anonymous competition on the Pokémon Showdown ranked ladder, where performance metrics are noisy and non-stationary. We address these challenges by releasing standardized baselines and datasets alongside a dedicated leaderboard for AI agents.

\subsection{Battling Evaluation Criteria}
\label{sec:track1:evaluation}

Battling Track agents are evaluated through direct competition against both community submissions and a diverse suite of state-of-the-art baselines maintained by our team. To avoid interfering with human players, all matches are conducted on a separate, modified Showdown server operated by the PokéAgent Challenge and configured specifically for AI benchmarking.

\vspace{-2mm}

\paragraph{Skill Rating Metrics.} Agents are ranked on a public leaderboard according to several metrics. We report the standard Showdown implementations of \textbf{Glicko-1}~\citep{glickman1999parameter} (an Elo variant incorporating uncertainty) and \textbf{GXE} (expected win probability against a randomly sampled opponent). However, these online metrics are designed for a large human player base with evolving skills and sparse matchups. In contrast, our agent pool is comparatively small, matchups are dense, and policies are fixed during evaluation. Our primary metric is based on a Bradley--Terry model~\citep{bradley1952rank} with bootstrapped uncertainty, fit over the full history of an agent’s battle results subject to a minimum sample size. We refer to this metric as the \textbf{Full-History Bradley--Terry} (\textbf{FH-BT}) rating to distinguish it from Showdown’s version of Elo, which is too noisy for our purposes. Appendix \ref{sec:track1-full-results} provides a comparison of alternative skill metrics.

\vspace{-2mm}

\paragraph{Rulesets.} Pokémon Showdown supports dozens of rulesets (``formats’’), but results here will focus on two that stress different AI capabilities: \textbf{Gen 1 OU} and \textbf{Gen 9 OU}. Gen 1 OU features greater effective hidden information and a more compact state space but yields smaller human demonstration datasets than Gen 9 OU. Our infrastructure currently supports three additional formats, with room to expand as performance saturates. Agents can play under two different time constraints: standard rules enforce faster-than-human play for efficient large-sample evaluation, while an ``Extended Timer'' variant provides nearly unlimited deliberation time for LLMs and test-time reasoning.

\subsection{Battling Baselines}

The PokéAgent Challenge is co-organized by the teams behind PokéChamp \cite{karten2025pok} and Metamon \cite{grigsby2025human}. While the Battling Track builds upon these leading LLM and RL approaches, the resources provided here have been heavily improved and standardized for this challenge. For clarity, we introduce these features as a unified framework, with novel improvements detailed in Appendix~\ref{sec:baseline-architectures}.

\paragraph{Demonstrations.} Showdown archives public battles spanning a decade of online play, and we organize an anonymized dataset of these files to protect player privacy. However, these ``replays'' are logged from a spectator’s perspective and do not reflect the private information available to each player at decision time. We release more than 4M RL trajectories generated by inferring private information and reconstructing the battle from each player’s perspective. The resulting dataset allows for flexible experimentation with alternative observation spaces, action spaces, and reward functions. While human demonstrations are invaluable for bootstrapping policies, competitive performance often requires the scale of self-play. We release all 18M trajectories used to train our strongest baselines and continue to expand this dataset with battles played on the PokéAgent Challenge server, including 100K community battles from our NeurIPS competition (Section~\ref{sec:neurips}).

\vspace{-2mm}

\paragraph{Sample Teams.} The combinatorial space of legal, competitively viable teams creates a substantial generalization challenge, as agents must perform across a vast range of initial conditions. Effective self-play training and evaluation demand diverse, realistic teams that mirror human trends. We release a dataset of 200K+ teams generated by inferring hidden information from human replays, alongside a curated collection of expert-validated teams sourced from community forums.

\begin{figure}[h!]
    \vspace{-2mm}
    \centering
    \makebox[\textwidth][c]{
        \includegraphics[width=1.02\textwidth]{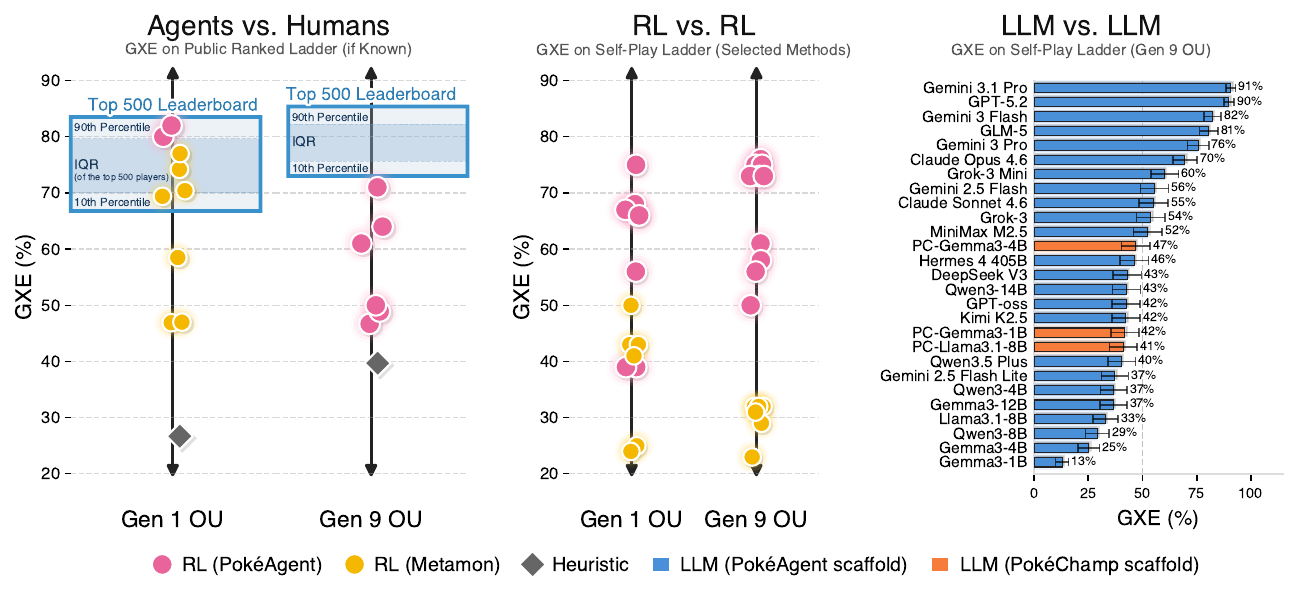}
    }
    \vspace{-7mm}
    \caption{\textbf{Baseline Performance.} \textbf{(Left) Agents vs. Humans}: Official ratings on the Showdown ladder. Statistics from the Top 500leaderboard are provided as a frame of reference for experienced human players. \textbf{(Center/Right) RL vs. RL and LLM vs. LLM}: GXE is measured relative only to methods within each plot. We differentiate between prior Metamon RL policies \cite{grigsby2025human} and baselines newly developed for this work; \texttt{PC-Llama3.1-8B} represents the original PokéChamp agent \cite{karten2025pok}.}
    \label{fig:track1_baselines}
\end{figure}

\vspace{-2mm}

\paragraph{LLM Baselines.} Although Pokémon knowledge appears in pretraining corpora, competitive gameplay is not an explicit optimization target of LLM training, making the application of that knowledge in competitive battles a genuine out-of-distribution test that extends recent LLM evaluations in Chess and Poker~\cite{kaggle2025gamearena} to an even more complex domain. We extend PokéChamp~\cite{karten2025pok} into a generalized harness framework for reasoning models, supporting both frontier API models (GPT, Claude, Gemini) and open-source models (Llama, Gemma, Qwen). The framework converts game state to structured text and provides configurable harness including depth-limited minimax search with LLM-based position evaluation. All LLM baselines use a harness; even small open-source models achieve meaningful performance with this support (Figure~\ref{fig:track1_baselines}). Default turn timers (60--90s) proved insufficient for LLM inference; the Extended Timer setting provides nearly unlimited deliberation time for fair evaluation of these methods. See Appendix~\ref{sec:baseline-architectures} for architecture details.

\vspace{-2mm}

\paragraph{RL Baselines.} In competitive domains, specialized systems often set the performance ceiling before general-purpose approaches reach parity. Pokémon provides a venue to study this gap, and we include strong RL baselines trained on large datasets of human demonstrations and self-play battles. We extend Metamon \cite{grigsby2025human} and release checkpoints from 30 agents that span the competitive skill ladder, ranging from compact RNNs to 200M-parameter Transformers. Our RL baselines provide high-quality reference points across a range of human skill levels, allowing researchers to benchmark their progress and explore compute-efficiency tradeoffs on accessible hardware.

Figure \ref{fig:track1_baselines} visualizes the relative strength of select RL and LLM agents alongside their performance against human players on Pokémon Showdown. Our baselines represent a substantial improvement over prior work \cite{karten2025pok, grigsby2025human} and span a broad performance range in both categories, providing researchers with diverse reference points to track progress as they iterate on new techniques. The strongest baselines are competitive with top human players, confirming the benchmark captures the strategic depth of high-level play, though our current upper bound remains less than superhuman.

\section{Long-Context Speedrunning Track}
\label{sec:track2}

Speedrunning provides a natural optimization objective for long-horizon planning: a clear metric (completion time), decomposable milestones for fine-grained credit assignment, and a task that demands the full stack of AI capabilities---visual perception, long-horizon planning, persistent memory, spatial navigation, and strategic combat---simultaneously. We formalize RPG gameplay as an episodic MDP $\mathcal{M} = (\mathcal{S}, \mathcal{A}, T, R, \gamma)$ where actions are button inputs, transitions are largely deterministic for navigation but stochastic for battles, and reward is $+1$ per milestone with $\gamma = 1$.

\subsection{Speedrunning Environment Design}

The evaluation environment runs the game server at a fixed frame rate. Agents receive visual frames and limited state information---party composition (species, levels), status conditions, and HP values---but puzzle states, dynamic obstacles, items, and movesets are not exposed, so that perception remains challenging (see Figure~\ref{fig:track2_interface} in Appendix~\ref{sec:environment-details}).

\subsection{Speedrunning Evaluation Criteria}

\begin{figure}[t]
    \centering
    \includegraphics[width=.98\linewidth]{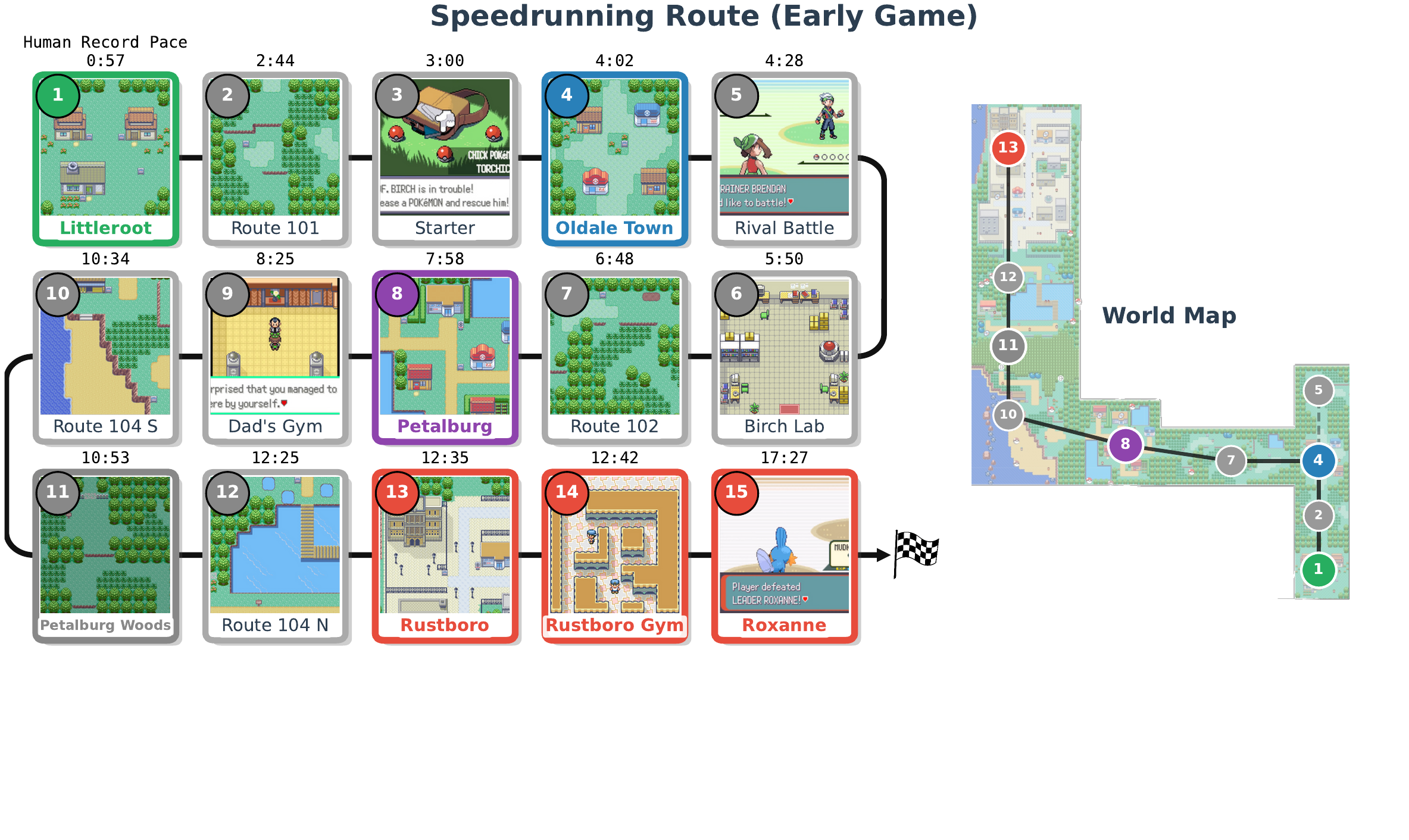}
    \vspace{-18mm}
    \caption{\textbf{Speedrunning Route (Early Game).} Milestones from \texttt{Littleroot Town} (1) to \texttt{Defeating Roxanne} (15), with game frames from each waypoint. The geographic overview (right) maps key locations. Although progression appears linear, the route requires substantial exploration and backtracking---agents must revisit earlier areas, navigate branching paths, and manage nonlinear dependencies between objectives. We provide splits from the human world record as an upper bound.}
    \vspace{-4mm}
    \label{fig:route}
\end{figure}

Agents are evaluated on \textbf{completion percentage} (progress through standardized milestones, illustrated in Figure~\ref{fig:route}) and \textbf{completion time} for agents achieving 100\%, with ties broken by action count. An \textit{action} is each discrete instance where the agent outputs button presses to the emulator. We scoped the initial evaluation to defeating the first gym leader (Roxanne). Even this early segment requires thousands of agent steps and millions of reasoning tokens, with agents maintaining coherent plans across extended context windows that accumulate over hours of real-time play. The task demands the full stack of AI capabilities---perception, memory, planning, navigation, and battle strategy---and requires context compaction to manage the thousands of reasoning steps involved. We scope to the first gym to enable rapid iteration on approaches at reasonable cost; the milestone framework naturally extends to the full game as agent performance saturates.

\vspace{-2mm}

\subsection{Speedrunning Baselines}

\paragraph{Human Baselines.} We scope evaluation to the first gym to enable participants to reasonably iterate on their approaches. Our top human speedrunner reached the first gym in 18 minutes, while average human players completed it in 1:22:05.

\vspace{-2mm}

\begin{figure}[t]
    \centering
    \includegraphics[width=.95\linewidth]{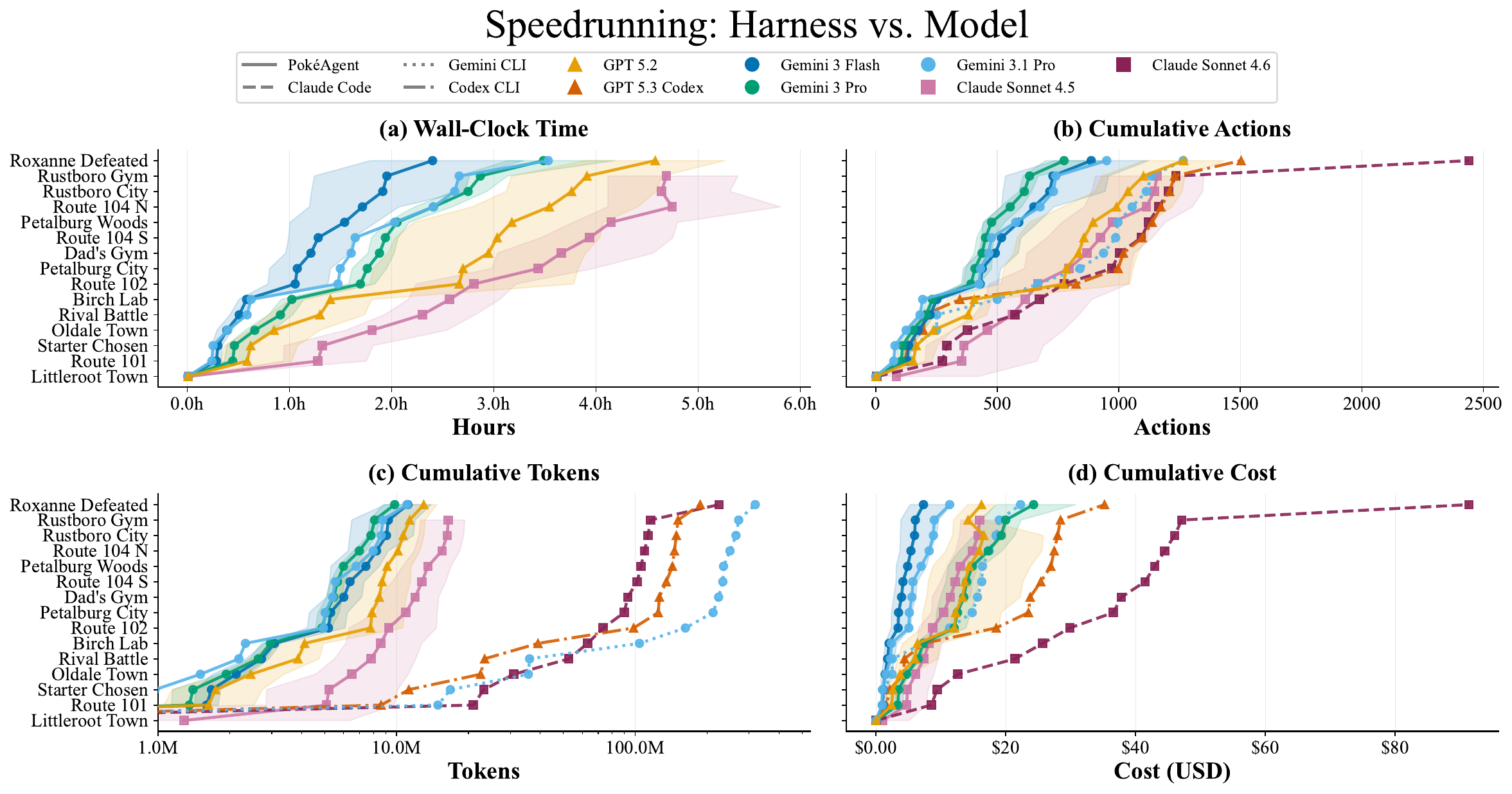}
    \caption{\textbf{Speedrunning Track Baseline Results.} Cumulative wall-clock time, actions, tokens, and cost at each milestone for five frontier models (mean $\pm$ min/max range across runs). Gemini 3 Flash completes the route fastest ($\sim$2:24 mean) but requires more actions than Gemini 3 Pro. Claude Sonnet 4.5 completes all milestones but with the highest variance and 3--4$\times$ the cost of Gemini variants. GPT-5.2 falls between the two families in both time and cost.}
    \label{fig:track2_results}
\end{figure}

\paragraph{Harness versus Model Capability.} A key challenge in evaluating LLM-based game agents is attribution: does performance stem from the underlying model or the surrounding harness (also called scaffold)? As discussed in Section~\ref{sec:intro}, prior efforts (Claude, Gemini, GPT playing Pokémon) conflated these factors. We disentangle them through a harness $\times$ model eval framework that analyzes systems along five dimensions---state representation ($S$), tools ($T$), memory ($M$), feedback ($F$), and fine-tuning ($\Phi$)---so that approaches can be compared on equal footing (Appendix \ref{sec:environment-details} Table~\ref{tab:system-comparison} and Figure~\ref{fig:track2_results}). Figure~\ref{fig:track2_results} compares our harness against common CLI-agent harnesses (Claude Code, Codex CLI, Gemini CLI) reveal that, while coding-agent architectures are  impressive out-of-the-box,  they nonetheless struggle to maintain coherence over the thousands of sequential decisions required for, and the non-linear exploration characteristic of RPG play.

\vspace{-2mm}

\paragraph{\pokeagent Baseline.} We release the first open-source multi-agent orchestration system for long-horizon RPG play. The system coordinates MCP tools (A* pathfinding, button inputs, knowledge retrieval) with specialized sub-agents for battle strategy, self-reflection, gym puzzles, and objective verification. A central orchestrator maintains a high-level route plan while dynamically dispatching sub-agents based on game context, with automatic context compaction to manage the thousands of reasoning steps required (see Appendix~\ref{sec:pokeagent-architecture} for architecture details). We evaluate five frontier models using the same harness (Figure~\ref{fig:track2_results}): Gemini 3 Flash achieves the fastest mean completion ($\sim$2:24), while Claude Sonnet 4.5 completes all milestones but with high variance (6:25--20:45 across runs). Even the best organizer baseline remains $\sim$1.8$\times$ slower than the average human (1:22:05).

\section{NeurIPS 2025 Competition}
\label{sec:neurips}

Our \href{https://pokeagent.github.io}{NeurIPS competition} provided an opportunity to grow the research community’s interest in Pokémon and validate our evaluation protocols. Participants competed for prizes across both our Battling and Speedrunning leaderboards. The competition, which ran from July to December 2025, grew an online community of 650+ members and generated 150+ submissions. Here, we summarize the competition's results and key takeaways; additional rules and details can be found in Appendix~\ref{sec:competition-retrospective}.

\subsection{Battling Track Results}

\begin{figure}[h!]
    \centering
    \includegraphics[width=\linewidth]{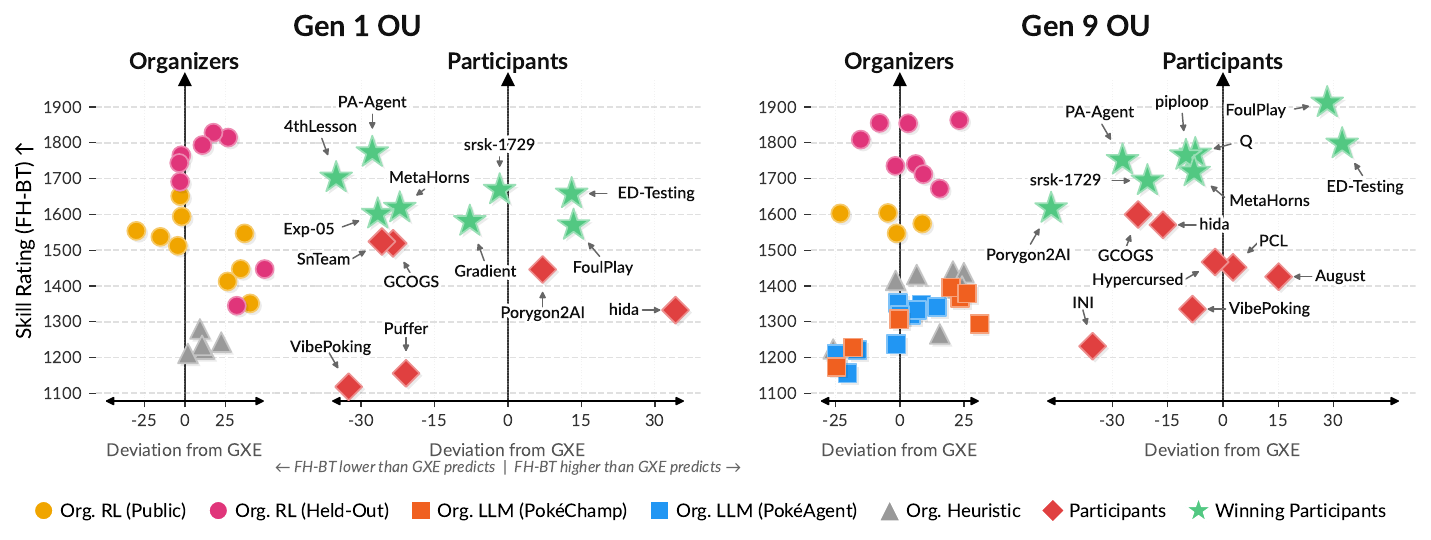}
    \vspace{-5mm}
    \caption{\textbf{NeurIPS Battling Leaderboard.} Organizer and Participant agent ratings are directly comparable. The x-axis measures disagreement between our primary rating metric and the linear trend established by GXE (a metric widely used on Showdown).}
    \label{fig:neurips_track1_main_text}
\end{figure}

Participants in the Battling Track submitted agents to our AI-focused Showdown leaderboard, where they competed against fellow entrants and organizer-hosted baselines. The PokéAgent Challenge offers substantial improvements over prior agents (Figure~\ref{fig:track1_baselines}); several of the strongest new baselines were kept private until the competition's conclusion to ensure that reaching the top of the leaderboard would require significant technical contributions. Figure~\ref{fig:neurips_track1_main_text} visualizes the final standings. The top 8 teams in both formats qualified for a head-to-head tournament bracket, where both \#1 seeds (\texttt{PA-Agent} in Gen 1 OU, \texttt{FoulPlay} in Gen 9 OU) were eventually victorious. Members of the winning and runner-up teams provide details of their final solutions in Appendix~\ref{sec:participant-methodologies}. Of the 16 qualifying spots, 13 were secured by teams extending our public RL baselines, while the remaining three were won by \texttt{Porygon2AI} (\#8 in Gen 9 OU) and \texttt{FoulPlay} \cite{FoulPlay} (\#1 in Gen 9 OU, \#8 in Gen 1 OU)---independent RL and search approaches, respectively. Appendix~\ref{sec:track1-full-results} provides further analysis of the results, including the final tournament, and a comparison of alternative rating schemes.

\vspace{-2mm}

\subsection{Speedrunning Track Results}

Of the 22 teams with valid submissions that submitted to the Speedrunning Track, 6 achieved 100\% completion (all 15 milestones). Figure~\ref{fig:track2_time_main} visualizes the final standings; full methodology descriptions appear in Appendix~\ref{sec:participant-methodologies}. The winner, Heatz, used Scripted Policy Distillation (SPD): an LLM decomposes the task into subgoals and generates scripted policies for each, which are then distilled into neural networks via imitation learning and refined with RL. The resulting agent completed the route in 40:13---nearly twice as fast as the second-place Hamburg PokéRunners, who used a pure RL approach (recurrent PPO with milestone-conditioned rewards). The remaining completions used LLM harness architectures with varying tool integrations (see Appendix~\ref{sec:track2-full-results}).

\begin{figure}[htb]
\centering
\includegraphics[width=\linewidth]{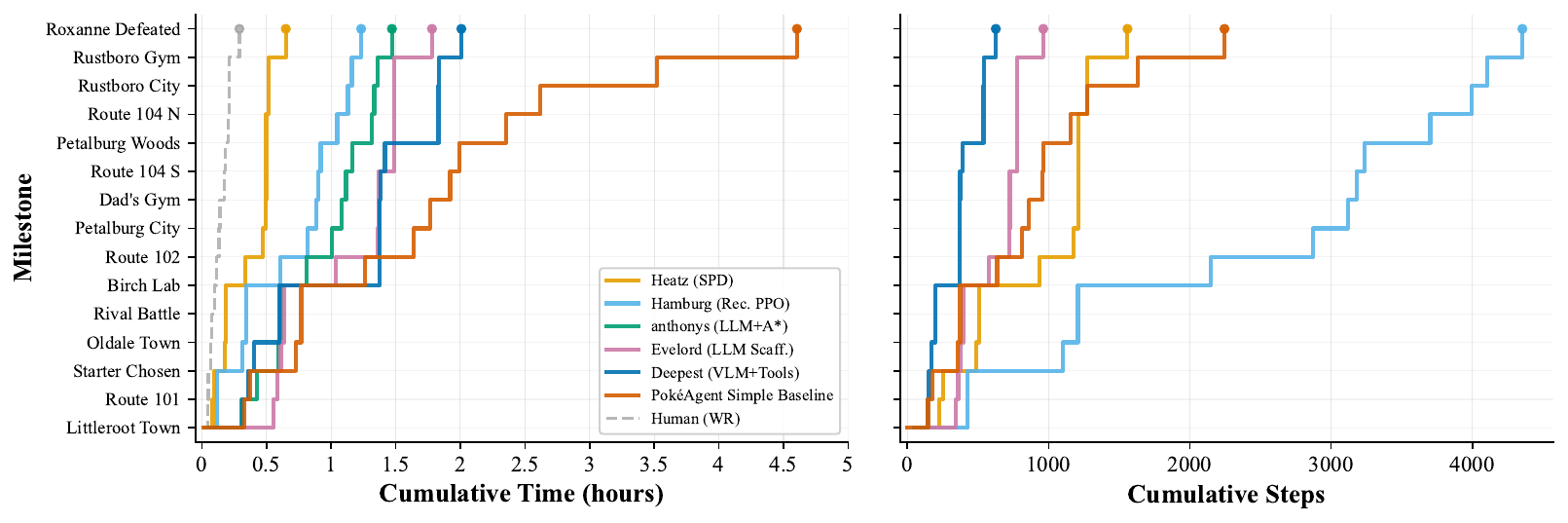}
\vspace{-5mm}
\caption{\textbf{NeurIPS Speedrunning Leaderboard.} Milestone progress vs.\ wall-clock time (left) and cumulative agent steps (right) for the six teams that completed all 15 milestones. Unlike most game benchmarks that pause between actions, our environment runs in real time---the game world continues while the agent reasons, making inference latency a first-class cost. We therefore report both wall-clock time (end-to-end throughput) and step count (sample efficiency). This distinction reveals that Deepest$^\dagger$ (Judge's Choice) completed in the fewest steps (649) despite ranking 5th by time, highlighting the tradeoff between inference speed and action efficiency.}
\label{fig:track2_time_main}
\end{figure}

Time-based rankings do not tell the full story. When measured by total actions rather than wall-clock time, Deepest---5th by time---completed with the fewest steps (649 vs.\ Heatz's 1,608), pointing to a tradeoff between inference speed and sample efficiency. Heatz's RL-distilled policy executes far faster per step, compensating for the higher step count.

Without a harness, raw frontier VLMs achieve effectively 0\% task completion on this track. Pokémon Emerald gameplay is out-of-distribution for these models, and raw model calls produce agents that wander aimlessly, repeat failed actions, or become stuck in dialogue loops. A harness---perception, memory, planning, and action modules---is not a marginal optimization but a prerequisite for any progress. Furthermore, not all harnesses are equal: common CLI-agent architectures (e.g., Claude Code, Codex CLI, Gemini CLI) fail to maintain coherence over the thousands of sequential decisions required, despite using the same underlying frontier models that succeed with our domain-specific harness (Figure~\ref{fig:track2_results}). Long-context autonomous embodied tasks appear to demand qualitatively different architectures than single-session coding workflows, particularly in the effective use of robust long-term memory abstractions coupled with detailed and consistently iterated plans specified at sufficient levels of granularity. In the absence of these abstractions, we find that CLI-agent architectures frequently make erroneous assumptions about their game progress and struggle to effectively localize their game progress within the broader continuity of milestone objectives in Pokémon Emerald. This gap likely extends to other OOD embodied domains requiring very long-context coherence.

\subsection{Cross-Track Insights}

\paragraph{Specialist methods outperform generalist LLMs.} Both tracks show a consistent pattern: RL and search methods outperform LLM approaches. In battling, the top participants all used RL or MCTS rather than LLM reasoning. In speedrunning, the top two finishers used RL-based methods. Heatz's 40:13 is more than $2\times$ faster than the best pure LLM harness approach (anthonys, 01:29:17). Put differently, Pokémon tasks require precise computation---probability estimation, opponent modeling, spatial planning---that current LLMs do not reliably perform from prompts alone.

\paragraph{LLMs as priors, RL as refinement.} Despite this gap, LLMs played a key role in the winning speedrunning approach. Heatz used an LLM to decompose the task and generate initial scripted policies, then distilled these into neural networks via RL. The LLM provided the prior (task structure and initial behavior) while RL provided refinement (faster execution and strategy discovery). This pattern---using LLMs for high-level reasoning and RL for low-level optimization---seems applicable beyond games.

\paragraph{Pokémon exposes reasoning failures that standard benchmarks miss.} In battles, weaker models exhibit ``panic behavior'' (also observed by \citep{gemini2p5report}): after a small tactical error, they compound mistakes rather than recovering, losing games they could still have won. Different model families fail in distinct ways---memory corruption cascades, goal oscillation, excessive plan commitment, and computational paralysis (see Appendix~\ref{sec:extended-discussion} and Figure~\ref{fig:cot_visualization}). These failure modes do not appear in coding or math benchmarks, where questions are independent. Pokémon's multi-turn, adversarial setting tests whether models recover from mistakes under pressure.

\paragraph{Pokémon battling is orthogonal to standard LLM benchmarks.} BenchPress~\citep{papailiopoulos2026benchpress} shows that an 83-model $\times$ 49-benchmark evaluation matrix is approximately rank-2: two latent dimensions explain $>$90\% of score variance, and 5 benchmarks suffice to predict the remaining 44 within $\sim$7 points. We added our Battling Track GXE scores for the 16 models that overlap with BenchPress. Pokémon breaks this low-rank structure: no existing benchmark correlates strongly with GXE (max Spearman $\rho = 0.77$; mean $|\rho| = 0.45$), and the rank-2 SVD that explains 91\% of standard benchmark variance captures only 27\% of GXE variance. Several models that score at frontier level on standard benchmarks collapse in battles, and vice versa. Competitive Pokémon appears to measure capabilities---strategic reasoning under partial observability and adversarial pressure---that are nearly orthogonal to what current evaluation suites capture.

\section{Conclusion: From Competition to Living Benchmark}
\label{sec:living-benchmark}

The \pokeagent Challenge transitions from the NeurIPS 2025 competition to a living benchmark available at \url{https://pokeagentchallenge.com}. The Battling Track maintains a live leaderboard on a dedicated Showdown server with all organizer baselines active, allowing new agents to be evaluated against the full history of submissions. The Speedrunning Track provides self-contained evaluation: researchers run agents locally against the standardized emulator and milestone framework, enabling reproducible comparison without server access. All datasets, baselines, and infrastructure are publicly released. The evaluation server and leaderboards are maintained by the organizing team with funding support as listed in the Acknowledgements; all code and data are hosted on GitHub and HuggingFace to ensure long-term availability independent of server infrastructure. Both tracks provide clear room to expand in order to maintain benchmark difficulty as performance saturates.

Large capability gaps remain. We highlight four open challenges: \textbf{(1) VLM-SLAM:} Speedrunning agents struggle with basic localization, action-distance estimation, and objective detection. Grounding VLM outputs to consistent spatial representations---analogous to classical SLAM but through language-vision interfaces---remains a bottleneck for RPG play. \textbf{(2) Closing the LLM--RL gap in battling:} Specialist RL agents far outpace harness LLM agents in competitive battles. Developing LLM agents that match RL performance, or hybrid approaches that combine RL's sample efficiency with LLM world knowledge, is an open problem. \textbf{(3) Full-game completion with open-source models:} Proprietary frontier models have completed Pokémon RPGs with heavy harness support, but no open-source model has done so. Achieving this would make long-horizon RPG evaluation accessible to more research groups. \textbf{(4) Approaching human speedrun times:} The best agent (Heatz, 40:13) is $2.2\times$ slower than human speedrunners. Closing this gap requires advances in navigation efficiency, obstacle avoidance, and objective sequencing---capabilities relevant to time-critical planning more broadly.

\section*{Acknowledgments}
We gratefully acknowledge funding from NeurIPS, IJCAI AI Journal, Google DeepMind, and compute support from Google Cloud Platform. This work is also supported by the NSF under Grant No. DGE-2444107. We thank all participating teams and community members for making this competition a success. We would also like to thank Aaron Traylor for his insights on competitive Pokémon and Caleb Frey for moderating our competition's online discussions. 

\bibliography{bib}

\newpage
\appendix

\startcontents[appendix]
\printcontents[appendix]{}{1}{\section*{Appendix Table of Contents}\setcounter{tocdepth}{2}}
\newpage

\section{Competition Retrospective}
\label{sec:competition-retrospective}

\subsection{Competition Setup}

\subsubsection{Timeline}

The \pokeagent Challenge ran from July 2025 through November 2025, culminating in presentations at NeurIPS 2025 in December. The timeline proceeded as follows: July launch with open ladder access, September hackathon for community building and technical talks, October qualifying rounds requiring 250+ games for statistical validity, and November finals featuring best-of-99 head-to-head matches. This structure allowed iterative improvement while maintaining evaluation integrity through held-out final assessments. The Speedrunning Track accepted submissions throughout with periodic leaderboard updates.

\subsubsection{Resources}

To support participants, we deployed several resources:
\begin{itemize}
    \item A dedicated custom Pokémon Showdown server hosting baseline agents and participant battles
    \item A RAG-powered Discord chatbot (``@pokeagent'') with a Professor Pokémon persona, using an embedding model and retrieval pipeline over organizer-curated documentation to answer participant questions
    \item Google Cloud Platform credits including Gemini API access
    \item A September hackathon with research talks streamed on YouTube
\end{itemize}

\subsection{Organizational Outcomes}

\paragraph{Participation Statistics}
The \pokeagent Challenge exceeded participation expectations:
\begin{itemize}
    \item \textbf{100+ active teams} with registered submissions across both tracks
    \item \textbf{650+ Discord community members} engaging in technical discussions
    \item \textbf{100K+ battles} on the competition Showdown server
    \item \textbf{22 valid Speedrunning Track submissions}, with 6 achieving 100\% completion
\end{itemize}

\paragraph{Dual Submission Tracks}
The dual-track structure successfully attracted participants from both the RL and LLM research communities. The Battling Track appealed to game AI and multi-agent learning researchers, while the Speedrunning Track attracted the long-context reasoning and language agent communities. Several teams competed in both tracks, developing unified architectures that could handle both strategic combat and exploration.

\paragraph{Cross-Platform Engagement}
The competition benefited significantly from the broader Pokémon AI zeitgeist of 2025. The Claude Plays Pokémon and Gemini Plays Pokémon communities provided a natural audience, and cross-promotion through Discord, Reddit (r/ClaudePlaysPokemon), and social media drove significant participation. The hackathon research talks, streamed on YouTube, attracted viewers from both academic and hobbyist communities.

\paragraph{Community Infrastructure}
Our RAG-powered ``@pokeagent'' Discord bot proved valuable for participant support, answering common questions about environment setup, baseline usage, and submission procedures using organizer-curated documentation. This reduced organizer burden while providing 24/7 assistance.

\paragraph{Timing Recommendations}
While NeurIPS encourages competitions to launch early, we found that peak participation and sustained engagement coincided with a 2--3 month window aligned with the Fall university semester (September--November). This timing particularly incentivized student involvement, as participants could integrate the competition into course projects and independent studies, creating a natural alignment between academic schedules and competition milestones.

\paragraph{Organizer Disclosure.} The organizing team includes the developers of PokéChamp and Metamon, which serve as baselines. To ensure fairness, organizer baselines were excluded from prize consideration, and several strong baselines were withheld during the competition to ensure that qualifying required genuine technical contribution beyond fine-tuning provided checkpoints. Of the 16 qualifying tournament spots, 13 were secured by teams that extended our public RL baselines, while 3 used independent approaches---reflecting that the released baselines provided an effective starting point rather than an unfair advantage. All evaluation was conducted on a shared server with identical conditions for all participants.

\section{Battling Track Competition Results}
\label{sec:track1-full-results}

The Battling Track competition was divided into a \textbf{Practice Stage}, \textbf{Qualifying Stage}, and \textbf{Tournament Stage}. In each phase, participating teams could compete in one or both of the Gen 1 OU and Gen 9 OU battling rulesets. The Practice and Qualifying Stages were most similar to the evaluation setup of the permanent PokéAgent Challenge leaderboard: teams connect their agents to our AI-focused Showdown server and compete in ranked battles against both their fellow participants and a large set of ``organizer baselines'' maintained by our team. Our baselines ensure accurate ratings by creating a shared set of opponents and maintaining battle activity when few participants are online. During the NeurIPS competition, organizer baselines also served as a form of held-out evaluation; some were released as part of the official starter kit, while others were kept private, so the only way to evaluate against them was to compete on the leaderboard.

\subsection{Practice Stage}
For most of the competition (July--October), participants were free to compete on the leaderboard to iterate on their solutions. Teams could create unlimited usernames to reset their ratings and evaluate new methods. A hackathon event in September reset the leaderboard and offered compute credit prizes to the top teams $36$ hours later, which were \texttt{ED-Testing} (Gen 9 OU) and \texttt{srsk-1729} (Gen 1 OU). In total, participants played about $75$k battles during the practice stage.

\subsection{Qualifying Stage}

In late October, our leaderboards were reset, and participation was restricted to one username per registered team. Teams competed to qualify for a spot in the Tournament Stage, where all teams would be guaranteed a cash prize. Qualifications were awarded to the top two teams by Elo and the next six highest by FH-BT, which is considered the more accurate metric (Appendix \ref{sec:rating-analysis}). This structure was a compromise that encouraged participation by allowing a late comeback, as Elo can be improved regardless of earlier results, and FH-BT required a minimum sample of $250$ battles, which could have been prohibitively expensive for some methods.

\begin{figure}[h!]
    \centering
    \includegraphics[width=.95\linewidth]{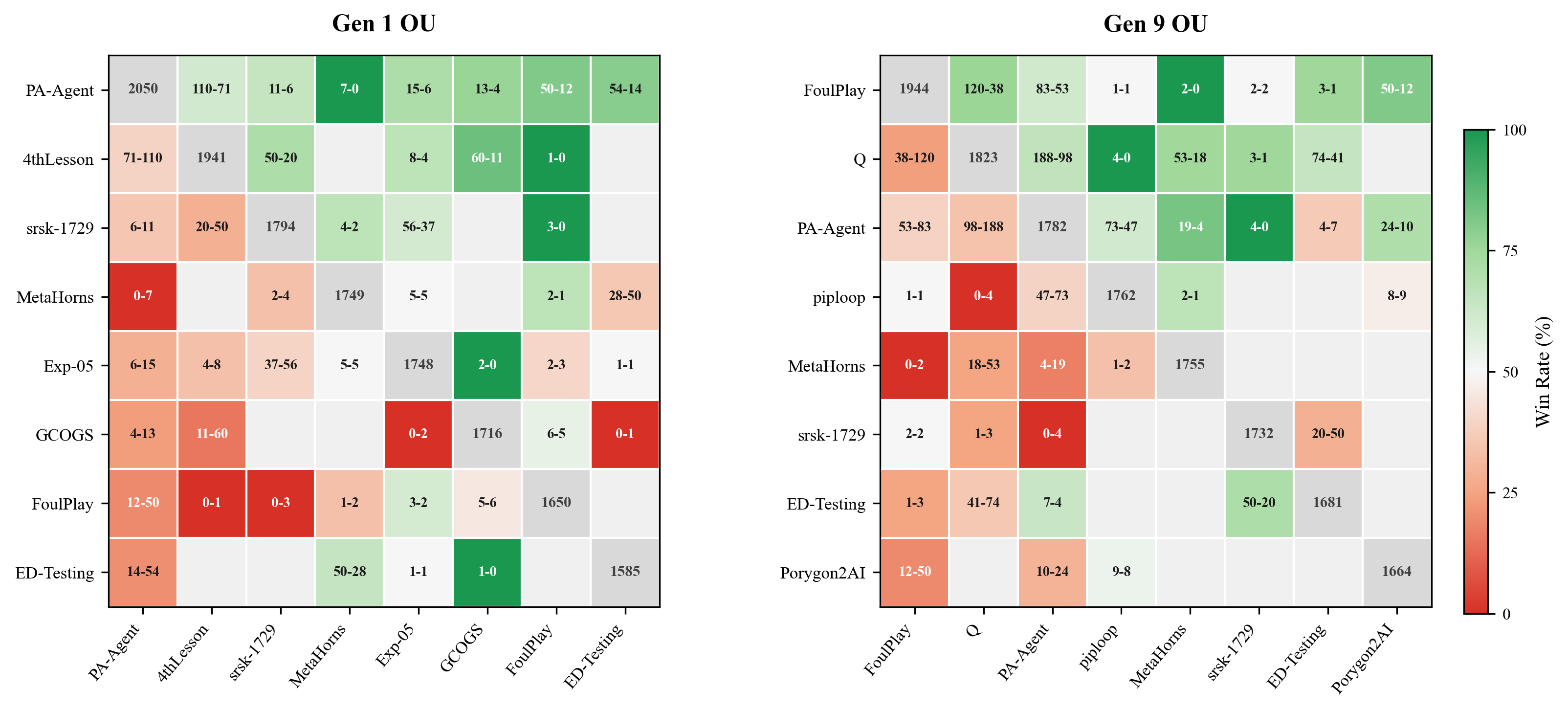}
    \caption{\textbf{Head-to-Head Win Rates Among Top Participants.} Heatmap matrices showing pairwise win rates for Gen 1 OU (left) and Gen 9 OU (right). Each cell shows the row player's win--loss record against the column player. Diagonal cells show each player's Elo rating. Players are ordered by Elo (highest at top). Empty cells indicate insufficient direct head-to-head matchups.}
    \label{fig:winrates}
\end{figure}

The main Qualifying Stage results were presented in Figure \ref{fig:neurips_track1_main_text}. Participants played a total of 35K battles over the two-week qualifying period, and we found strong alignment across all of our skill rating metrics; Figure \ref{fig:track1_qualifying_scatter_appendix} visualizes the leaderboard across several alternatives. Figure~\ref{fig:winrates} presents head-to-head win rates among the top non-baseline participants. 

Our set of organizer baselines helps give the participant results context by grounding them against methods whose training details and performance relative to human players are known. The distinction between public and private organizer baselines created a clear separation point on both leaderboards: many participants clustered slightly above the best public baseline, whereas participants competitive with the private baselines separated from the field and qualified for the tournament. In Gen 9 OU, \texttt{FoulPlay} outperformed private baselines by all metrics, while \texttt{Q}, \texttt{PA-Agent}, and \texttt{piploop} formed a chase pack situated between the best public and private baselines. In Gen 1 OU, \texttt{4thLesson} separated from the public baselines while \texttt{PA-Agent} fell just short of the best private baselines and placed first among participants by a wide margin. 

As discussed in Section~\ref{sec:neurips}, all top-performing submissions employed RL or search-based methods rather than pure LLM approaches. Our baselines revealed systematic LLM reasoning failures---including ``panic behavior'' and other failure modes---discussed in Section~\ref{sec:neurips} and Appendix~\ref{sec:extended-discussion}. Additionally, LLM agent performance on the competition ladder was deflated by time pressure: agents frequently switched to less sophisticated fallback methods when low on time, or lost on time entirely. We have since introduced a separate long-timer leaderboard to isolate reasoning ability from inference speed (Section~\ref{sec:track1:evaluation}).

\subsection{Tournament Stage}

\begin{figure}[h!]
    \centering
    \includegraphics[width=0.98\linewidth]{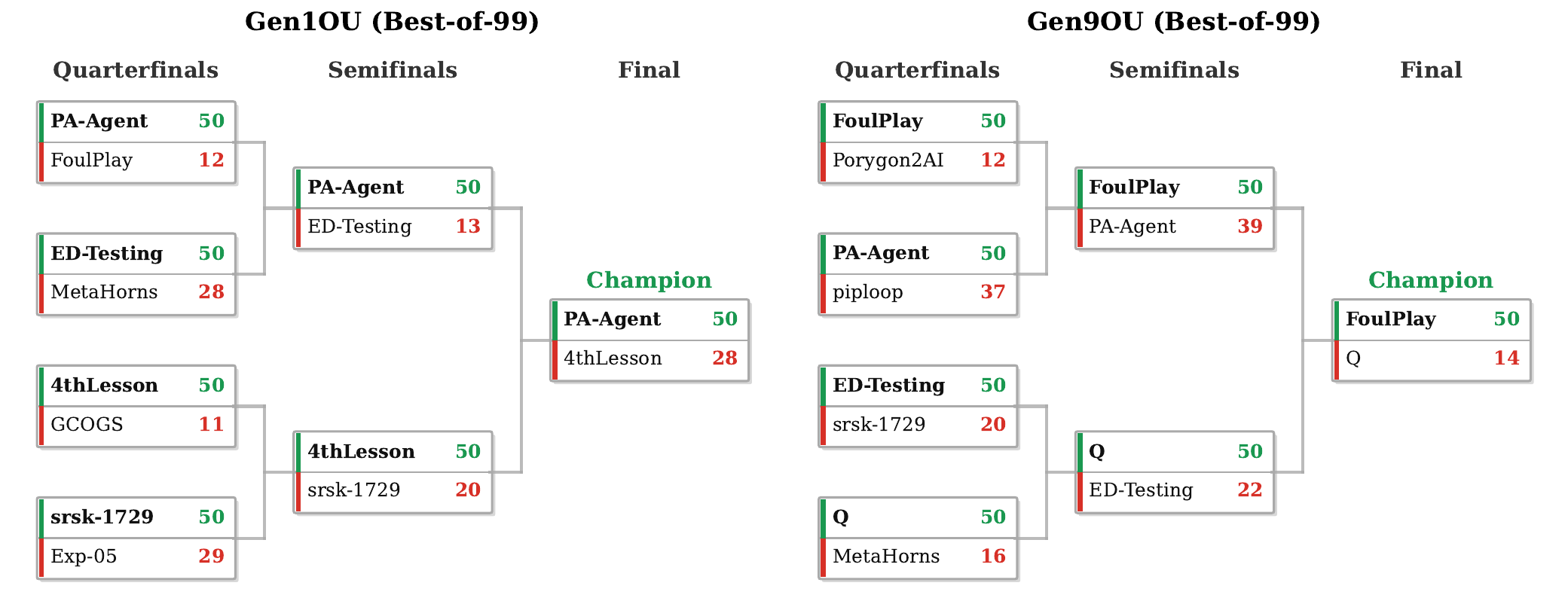}
    \caption{\textbf{Tournament Brackets.} Best-of-99 single-elimination brackets for Gen 1 OU (left) and Gen 9 OU (right). Numbers indicate games won by each team; green highlighting indicates winners.}
    \label{fig:tournament}
\end{figure}

The competition concluded with a head-to-head single-elimination tournament among the qualifying teams, seeded by Qualifying Stage leaderboard position. Teams played their designated opponent in a best-of-99 battle match (first to 50 wins). Figure~\ref{fig:tournament} shows the tournament brackets for both formats. \texttt{PA-Agent} dominated Gen 1, defeating \texttt{4thLesson} 50--28 in the finals. In Gen 9, \texttt{FoulPlay} swept through the competition, defeating \texttt{Q} 50--14 in the finals after a close match with \texttt{PA-Agent} in the semifinals. Both tournament winners were ranked \#1 on the Qualifying Stage leaderboard and were therefore the \#1 seeds. Appendix \ref{sec:participant-methodologies} provides technical details of the first and second place methods as written by the teams themselves.

\subsection{Rating System Analysis}
\label{sec:rating-analysis}

\begin{figure}[h]
    \centering
    \includegraphics[width=0.95\linewidth]{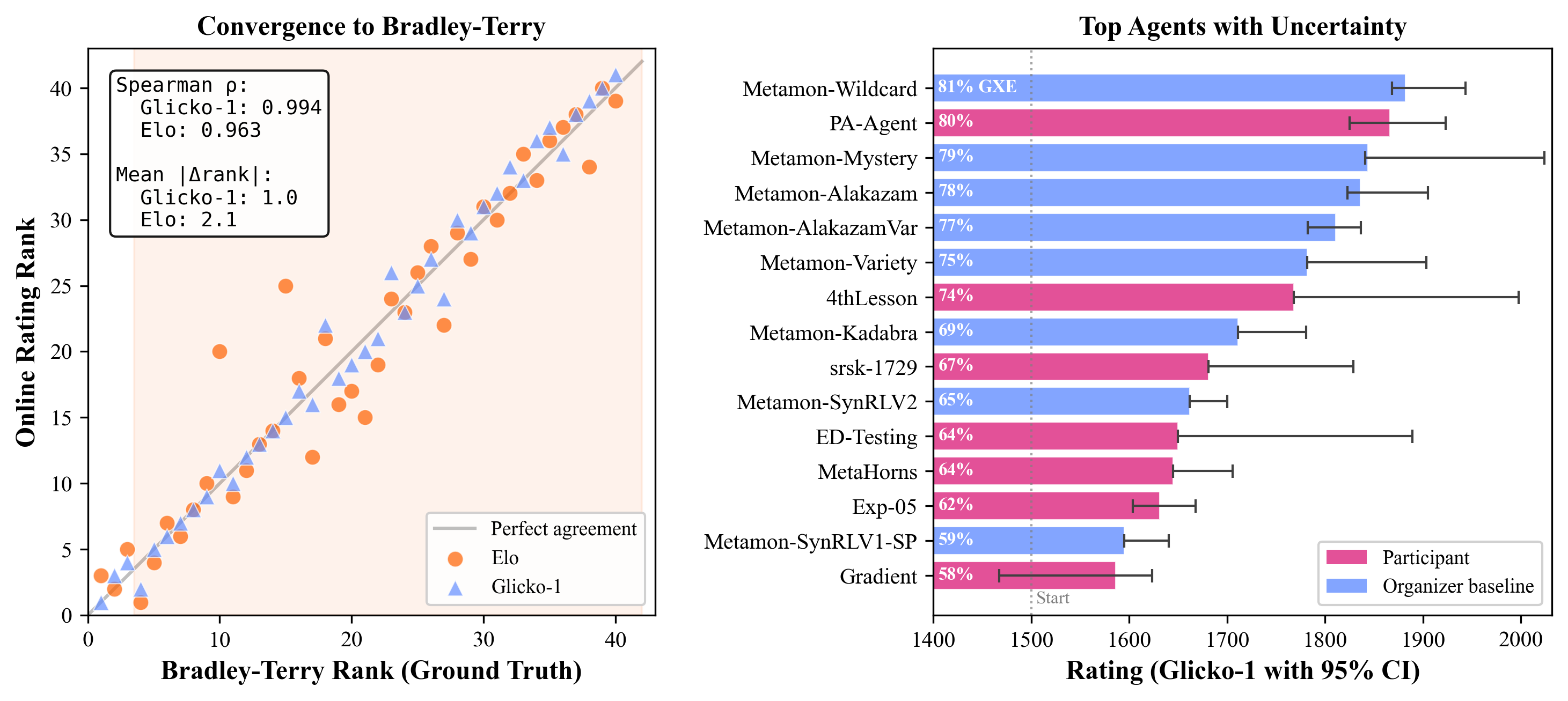}
    \caption{\textbf{Rating System Comparison Across 1.6M+ Matches.} Four ranking methods applied to Battling Track data. Left: Rank correlation between methods, showing Glicko-1 converges to Bradley--Terry (batch MLE) while Elo diverges for mid-ranked agents. Right: Rating trajectories with uncertainty bands for selected agents, demonstrating Glicko-1's uncertainty quantification. Top-3 agents (shaded) show consistent rankings across methods; ranks 4+ exhibit systematic disagreement, particularly between Elo and Bradley--Terry.
    }
    \label{fig:rating_systems}
\end{figure}

\begin{figure}
    \centering
    \includegraphics[width=\linewidth]{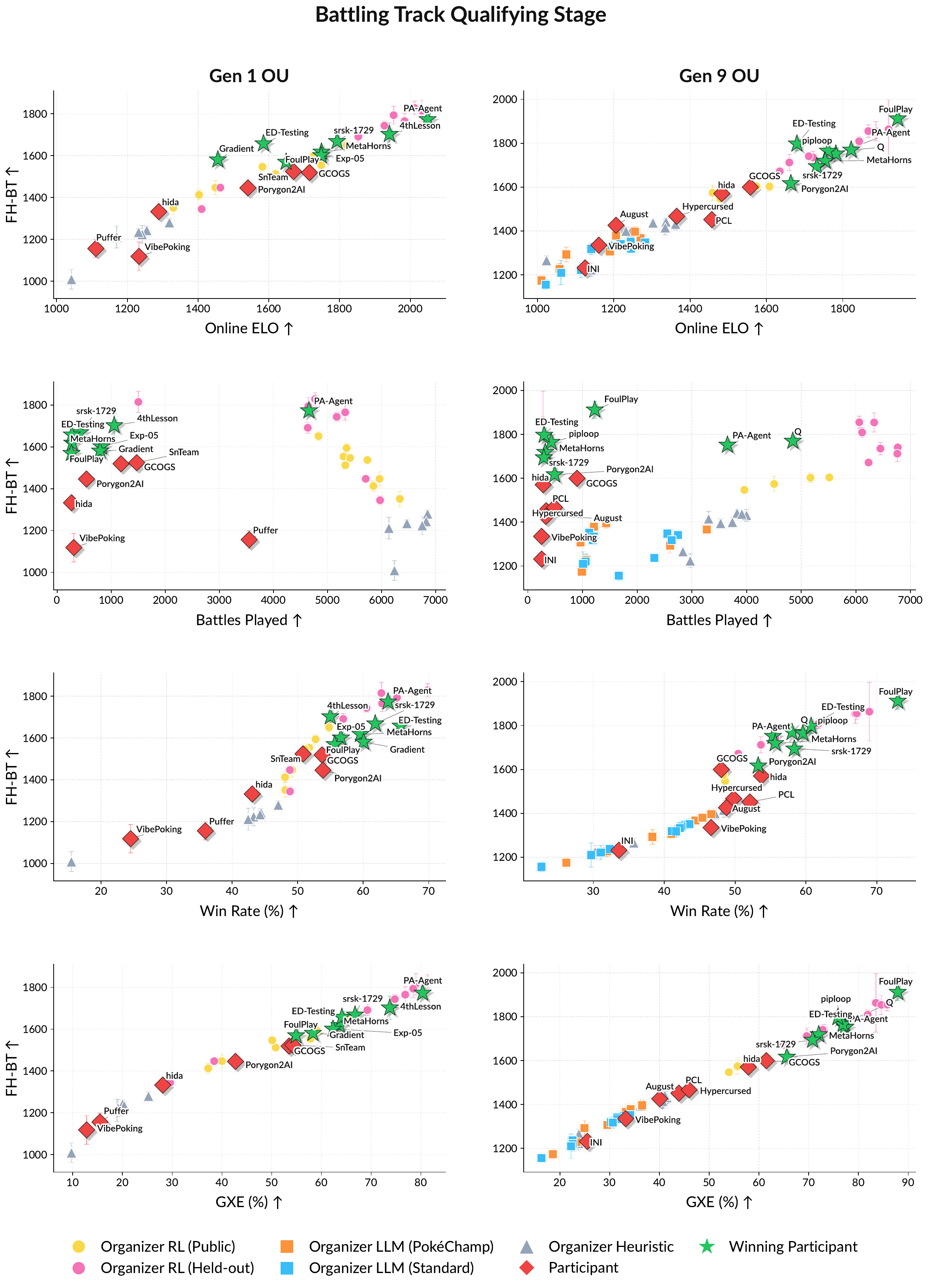}
    \vspace{2mm}
    \caption{\textbf{Battling Track Qualifying Metrics.} Qualifying stage results according to alternative skill metrics (Showdown's Elo, GXE, Win Rate) and sample size (Battles Played). }
    \label{fig:track1_qualifying_scatter_appendix}
\end{figure}

Most LLM evaluation arenas use Bradley--Terry models (batch MLE) without reporting uncertainty---a potentially misleading practice. We stress-tested four ranking systems (Bradley--Terry, Elo, Glicko-1, and GXE) across 1.6M+ agent matches (Figure~\ref{fig:rating_systems}). Top-3 agents converged across all methods (with 250+ games each), but ranks 4+ showed systematic disagreement---Elo diverged from Bradley--Terry even when error bars did not overlap. Glicko-1 offered the best tradeoff: online updates for real-time leaderboards with convergence guarantees matching batch MLE. We prefer Glicko-1 over raw Elo for agent evaluation, particularly when sample sizes vary across agents. Figure \ref{fig:track1_qualifying_scatter_appendix} compares the Battling Track competition results according to alternative skill metrics (Showdown's Elo, GXE, Win Rate) and sample size (Battles Played).

\subsection{Judge's Choice Awards:}

The organizing committee awarded additional prizes based on participant technical writeups. These Judge's Choice awards were intended to reward novel directions and significant departures from the starter kit baselines regardless of final results. The winners were:
\begin{itemize}
    \item \textbf{\texttt{Porygon2AI}}: Recognized for innovative league training methodology inspired by AlphaStar's~\citep{vinyals2019grandmaster} approach to diverse opponent modeling
    \item \textbf{\texttt{August}}: Best pure LLM method without learned components, demonstrating effective chain-of-thought reasoning for move selection
\end{itemize}

\section{Speedrunning Track: Full Competition Results}
\label{sec:track2-full-results}

The Speedrunning Track challenged participants to complete Pokémon Emerald as quickly as possible (see Appendix~\ref{sec:participant-methodologies} for detailed descriptions of participant methods). Of 22 teams that submitted, 6 achieved 100\% completion. Figure~\ref{fig:track2_time_main} presents the final standings, showing both wall-clock time and step-count views. Heatz won with a 40:13 completion time using Scripted Policy Distillation (SPD); see Section~\ref{sec:neurips} for a discussion of the winning approach and the time-vs-efficiency tradeoff across teams.

As discussed in Section~\ref{sec:neurips}, a harness is a prerequisite for any OOD performance---not a marginal optimization. The successful harness architectures decompose into four components:
\begin{itemize}
    \item \textbf{Perception}: Vision-to-text translation for game state understanding
    \item \textbf{Memory}: Maintaining coherent state across thousands of timesteps
    \item \textbf{Planning}: High-level goal decomposition and route optimization
    \item \textbf{Action}: Low-level control for navigation and battle execution
\end{itemize}
Benchmark results therefore reflect the joint capability of model + harness, and differences between models cannot be attributed to ``raw'' model capability alone.

\textbf{Judge's Choice:} Deepest was recognized for achieving completion with the fewest total actions, demonstrating efficient reasoning despite a slower wall-clock time.

\section{Baseline Architecture Details}
\label{sec:baseline-architectures}

\subsection{Battling Track Baselines}

PokéChamp \cite{karten2025pok} and Metamon \cite{grigsby2025human} form the foundation for our Battling Track baselines. Both projects were significantly upgraded to improve their competitive performance and serve as extensible starter points for future work. All of these changes are open-source and detailed information is available via the \href{https://pokeagentchallenge.com}{PokéAgent Challenge Resources Page}. This section provides a high-level introduction and summarizes key improvements.

\subsubsection{PokéChamp}

PokéChamp combines large language models with minimax search for competitive battling. The system converts game state to text via a game engine interface, then uses an LLM to evaluate positions through an approximate state transition heuristic. This enables strategic lookahead without requiring explicit reward signals or training data.

\begin{figure}[h]
    \centering
    \includegraphics[width=.95\linewidth]{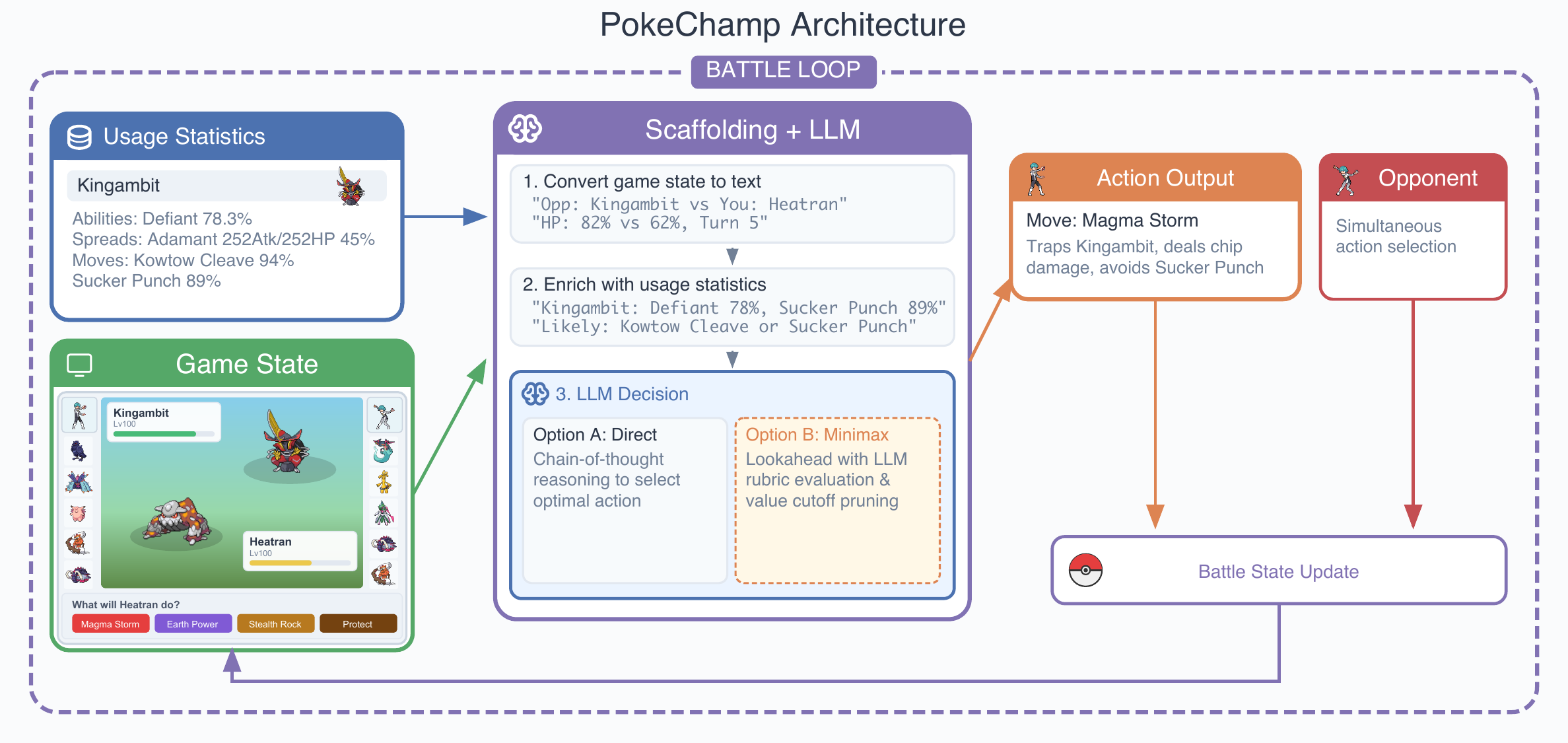}
    \caption{\textbf{PokeChamp Architecture.} Within the battle loop, two data sources feed into the LLM pipeline: (1)~historical usage statistics from Smogon (left top), providing probability distributions over abilities, EV spreads, and moves for each species; and (2)~the current game state from Pokémon Showdown (left bottom). The Text Conversion + LLM module (center) converts game state to structured text, enriches it with usage statistics for opponent prediction, and performs LLM inference to select actions. The chosen move executes and updates the game state for the next turn.}
    \label{fig:pokechamp}
\end{figure}

\subsubsection{Metamon}

\begin{figure}[h!]
    \centering
    \includegraphics[width=.80\linewidth]{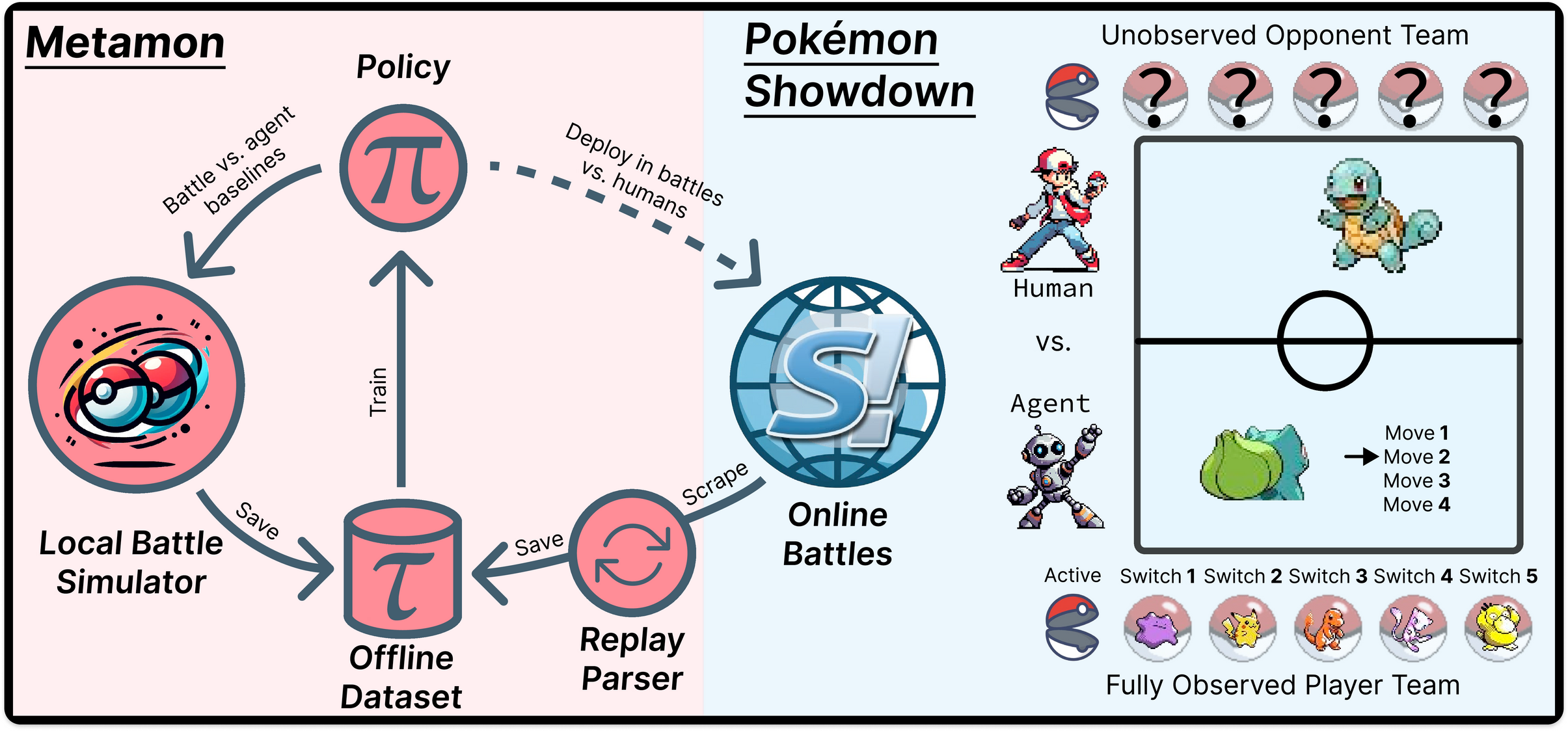}
    \caption{\textbf{Metamon Training Pipeline.} Metamon trains RL agents on large datasets of self-play battles and human demonstrations ``reconstructed'' from replays gathered from Pokémon Showdown. Figure reproduced from \citet{grigsby2025human}.}
    \label{fig:metamon}
\end{figure}

Metamon converts Showdown's replay archives into an opportunity to study offline RL \cite{levine2020offline} at scale in a complex domain. Its baselines embrace the complexity and partial observability of Pokémon by training Transformer policies with model-free off-policy RL updates \cite{grigsbyamago}. Further improvements beyond the human demonstration dataset are driven by a large-scale ``replay across experiments'' \cite{tirumalareplay} loop in which the project continuously expands a central repository of self-play trajectories and policies (Figure \ref{fig:metamon}). As of the launch of the PokéAgent Challenge, the training dataset exceeds $20$M battles and has produced $30$ meaningfully distinct policies at various skill levels. Metamon training runs are expensive by academic standards and each iteration may have several axes of improvement (datasets, architectures, RL details). As a result, its agent releases are given somewhat arbitrary names (LLM-style) that are excluded from results like Figures \ref{fig:track1_baselines} and \ref{fig:neurips_track1_main_text} but are available in PokéAgent Challenge resources. Table \ref{tab:metamon_baselines} defines a short list that are relevant to the NeurIPS competition results (Appendix \ref{sec:track1-full-results}) or referenced by participant writeups in (Appendix \ref{sec:participant-methodologies}).

\begin{table}[htbp]
\centering
\renewcommand{\arraystretch}{1.3} 
\resizebox{.95\textwidth}{!}{
\begin{tabular}{l c c p{10cm} c c c c c}
\toprule
\multirow{2}{*}{\textbf{Model}} & \multirow{2}{*}{\textbf{Size}} & \multirow{2}{*}{\textbf{Date}} & \multirow{2}{*}{\textbf{Description}} & \multicolumn{5}{c}{\textbf{Human Ladder Ratings (GXE)}} \\
\cmidrule(lr){5-9}
& & & & \textbf{G1} & \textbf{G2} & \textbf{G3} & \textbf{G4} & \textbf{G9} \\
\midrule
\texttt{SyntheticRLV2} & 200M & Sep 2024 & The best (public) Gen 1 OU policy during the NeurIPS competition, and therefore the basis of many of the qualifying submissions (Appendix \ref{sec:participant-methodologies}). & 77\% & 68\% & 64\% & 66\% & -- \\
\addlinespace
\texttt{Abra} & 57M & Jul 2025 & The best (public) Gen 9 OU policy during the NeurIPS competition, and therefore the basis of many of the qualifying submissions (Appendix \ref{sec:participant-methodologies}). & -- & -- & -- & -- & 50\% \\
\addlinespace
\texttt{Kadabra3} & 57M & Sep 2025 & The best policy to participate in the NeurIPS competition as an organizer baseline (rank \#1 in the Gen 1 OU qualifier and \#2 in Gen 9 OU). & 80\% & -- & -- & -- & 64\% \\
\addlinespace
\texttt{Kakuna} & 142M & Dec 2025 & The best metamon model as of the launch of the PokéAgent Challenge (with results appearing in Fig. \ref{fig:track1_baselines}). & 82\% & 70\% & 63\% & 64\% & 71\% \\
\bottomrule
\end{tabular}
}
\vspace{1mm}
\caption{Metamon baselines referenced in Appendix results, and their estimated performance against human players. ``G1'' $\rightarrow$ ``Gen 1 OU''.}
\label{tab:metamon_baselines}
\end{table}

The most notable improvement made to Metamon for the PokéAgent Challenge was its expansion to support the Gen 9 OU ruleset, which created the opportunity for RL and LLM baselines to compete directly. Gen 9 OU has a significantly larger state space (Appendix \ref{sec:state-space-derivation}) than the rulesets from the original work, and it was an open question whether model-free RL would generalize to this more complex game. Today, Metamon is at least as competitive against human players in Gen 9 OU as earlier generations (Table \ref{tab:metamon_baselines}), though its performance in Gen 1 OU remains an outlier. 

\subsection{Speedrunning Track Baselines}
\label{sec:pokeagent-architecture}

The \pokeagent starter kit implements a \textbf{multi-agent orchestration system} that successfully completed Pokémon Emerald. The architecture distinguishes between two types of capabilities: \textit{MCP tools} for low-level game interaction and \textit{sub-agents} for high-level reasoning. A central orchestrator coordinates these components, dispatching to specialized sub-agents based on game context while using MCP tools for direct game control.

\begin{figure}[h]
    \centering
    \includegraphics[width=\linewidth]{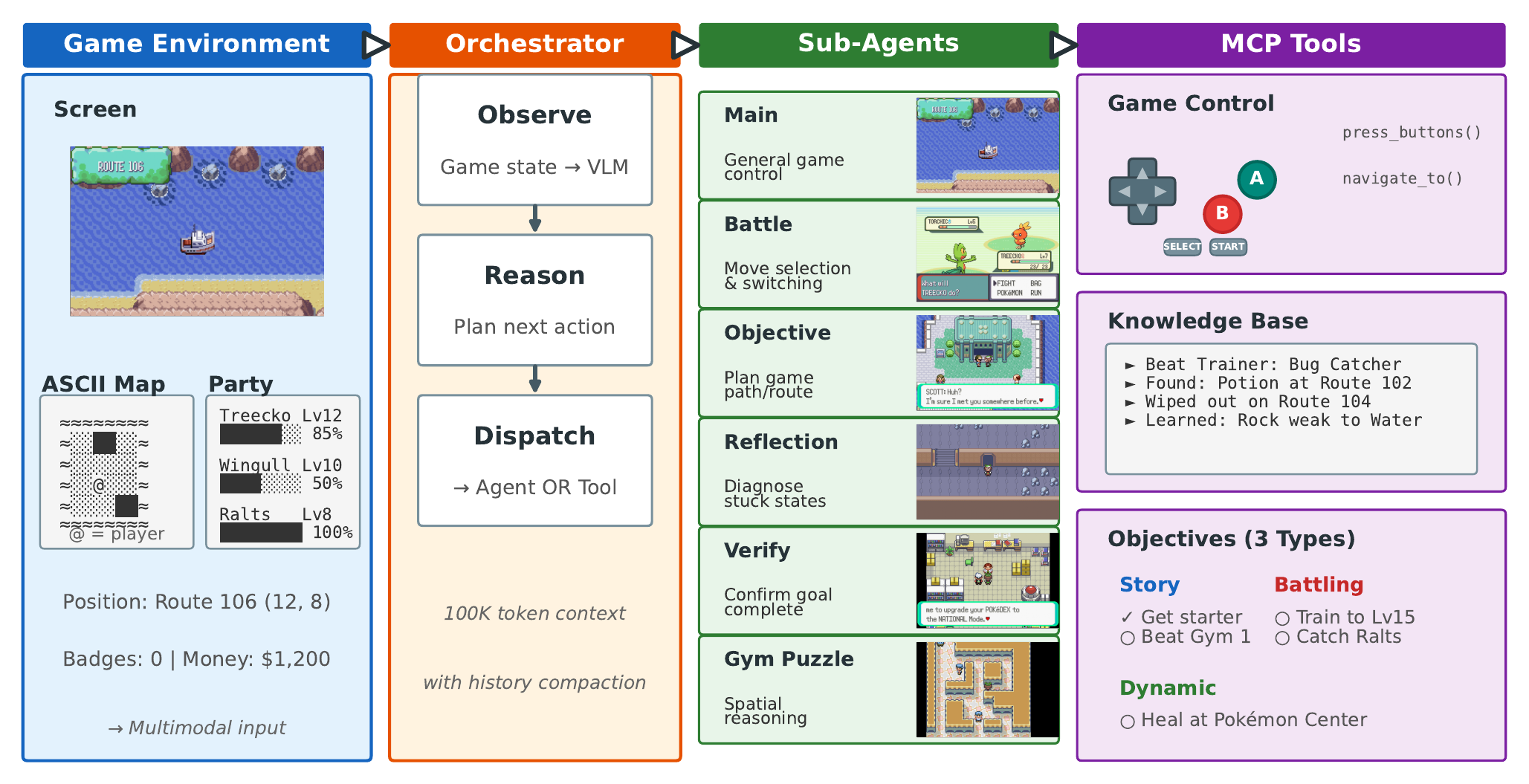}
    \caption{\textbf{\pokeagent Multi-Agent Architecture.} The orchestrator coordinates sub-agents and MCP tools to play the game. Data flows left-to-right from the GBA emulator (game frames, parsed state) through the Orchestrator (context management, sub-agent dispatch, stuck detection), which invokes either MCP tools (pathfinding, button inputs, knowledge retrieval) for direct game control or specialized sub-agents (battle agent, reflection agent, verification agent, gym puzzle agent) for complex reasoning tasks. A feedback loop returns sub-agent outputs and tool results to inform subsequent decisions.}
    \label{fig:pokeagent_arch}
\end{figure}

\paragraph{Orchestrator.} The flagship harness implements a central orchestrator that coordinates all sub-agents and tool calls. Upon initialization, the orchestrator invokes a \textit{planning sub-agent} to generate a high-level route covering story progression, team building objectives (training, catching, item acquisition), and resource management (Pokémon Center visits). This loose plan maintains long-term coherence while allowing ad-hoc objective creation during gameplay. The orchestrator dispatches to specialized sub-agents based on context: the \textit{battle sub-agent} handles combat decisions, the \textit{reflection sub-agent} analyzes stuck states, and the \textit{verification sub-agent} checks objective completion to prevent premature advancement.

The system exposes two categories of capabilities to the orchestrator:

\paragraph{MCP Tools.} Low-level game interaction via HTTP endpoints:
\begin{itemize}
    \item \textbf{Game Control}: \texttt{get\_game\_state}, \texttt{press\_buttons}, and \texttt{navigate\_to} (A* pathfinding with variance options for obstacle handling)
    \item \textbf{Knowledge Retrieval}: Persistent memory system storing discoveries (locations, NPCs, items, strategies) with importance-weighted retrieval.
    \item \textbf{Objective Management}: Three parallel objective sequences---\textit{story} (main narrative), \textit{battling} (team building), and \textit{dynamics} (agent-created adaptive goals)---enabling balanced progress across game dimensions
\end{itemize}

\paragraph{Sub-Agents.} Specialized reasoning modules invoked by the orchestrator:
\begin{itemize}
    \item \textbf{Battle Sub-Agent}: Handles combat decisions including move selection, switching, and item usage based on type matchups and team state
    \item \textbf{Reflection Sub-Agent}: Analyzes stuck states by comparing current situation against ground truth sources (porymap data, knowledge base) to diagnose navigation failures
    \item \textbf{Verification Sub-Agent}: Independently checks whether objectives are truly complete, preventing the orchestrator from advancing when the main agent incorrectly believes a task is finished
    \item \textbf{Gym Puzzle Sub-Agent}: Specialized reasoning for gym-specific puzzles requiring spatial planning (e.g., Mauville's electric barriers, Lavaridge's floor tiles)
\end{itemize}

Interestingly, wiki access via the knowledge retrieval tool sometimes degraded performance---the agent would retrieve contradictory information from different sources or misapply advice for different game versions, highlighting the challenge of grounding external knowledge. Thus, we did not grant the agent access to searching the web or Pokémon specific wikis for our baselines.

\paragraph{Context Management.} The agent maintains conversation history with automatic compaction when exceeding history time-step limit (20+ turns typically), preserving only LLM responses and action summaries while discarding redundant game state. The knowledge base summary maintains persistent memory. This enables coherent behavior across thousands of timesteps without context overflow.

\paragraph{VLM Backend Support.} Both the orchestrator and all sub-agents share a unified VLM interface supporting multiple backends (Gemini, OpenAI, OpenRouter), with action speed control (fast/normal/slow) for different gameplay situations---rapid button presses for dialogue advancement versus deliberate inputs for critical menu navigation.

\section{Participant Methodologies}
\label{sec:participant-methodologies}

This section presents methodology summaries contributed by top-performing teams. These descriptions are provided largely verbatim to preserve the participants' voices and technical details, but we note that their writing generally assumes familiarity with Pokémon terminology and PokéAgent Challenge resources (baseline names, dataset names, etc.).

\subsection{Battling Track: Competitive Battling}

\subsubsection{PA-Agent (Gen 1 OU Champion)}
\label{sec:pa-agent}

\textbf{Team:} Xianwei Shi, Kunsheng Zhou, Dongyu Liu, Wenli Zhang \\
\textbf{Affiliation:} PokéAgent Challenge Gen 1 OU Champion

PA-Agent addresses Pokémon battle challenges (massive team composition space and incomplete information) by building on the Metamon offline reinforcement learning (RL) framework~\citep{grigsby2025human}, integrating Transformer-based decision-making, iterative training, and tournament-driven team selection. The approach prioritizes efficiency and adaptability to stochastic, partial-observability scenarios.

The agent adopts a modular architecture with two core components: a Battle Decision Module and a Team Optimization Module. The Battle Decision Module uses a Transformer backbone to process hybrid inputs (87 textual tokens and 48 numerical features) for end-to-end action prediction (move selection/switching), enhancing attention to recent battle events to mitigate information overload. Key algorithmic innovations include iterative offline RL with dynamic data weighting---bootstrapping from human replays provided by Metamon, then refining via 6 rounds of inter-model battle data, gradually reducing human data proportion from 100\% to 10\% to avoid low-quality decision interference. For team composition, a tournament selection mechanism narrows the huge state space by evaluating candidate teams (from Smogon, Metamon, and community sources) against 50+ lineups, selecting those with $>$60\% win rates.

In the Pok\'eAgent Challenge, PA-Agent achieved competitive performance against top submissions, demonstrating the effectiveness of iterative offline RL refinement and tournament-based team selection. The model's GXE in Gen 1 OU qualifying reached 80.35\%, with particular strength in handling diverse opponent strategies and partial observability.

\subsubsection{FoulPlay (Gen 9 OU Champion)}
\label{sec:foulplay}

\textbf{Author:} Patrick Mariglia~\citep{FoulPlay} \\
\textbf{Affiliation:} PokéAgent Challenge Gen 9 OU Champion

Foul Play is a competitive Pokémon battle bot that employs root-parallelized Monte Carlo Tree Search (MCTS) with the Decoupled Upper Confidence Bound applied to Trees (DUCT) formula to handle simultaneous move selection. The bot uses a custom battle engine called \texttt{poke-engine}, written in Rust for performance, which addresses the computational complexity inherent in competitive Pokémon. Rather than exhaustively exploring all possible game states (which is intractable due to the combinatorial explosion from damage rolls, critical hits, and secondary effects), \texttt{poke-engine} employs damage roll grouping. This technique clusters damage outcomes by their practical impact---primarily whether they result in a knockout---reducing branching while preserving strategically relevant information. The engine generates state instructions rather than copying entire game states, enabling efficient tree traversal to depths of 10 or more turns for promising lines while pruning uninteresting branches early.

The transition from expectiminimax to MCTS was motivated by depth limitations: the earlier approach would exhaustively search every branch of the game tree, limiting depth to approximately 5 turns. MCTS achieves 10+ turn depth on promising lines while exploring unpromising branches to only 2--3 turns. Rather than relying on rollouts, search is guided by a custom evaluation function.

Set prediction is critical to Foul Play's performance, particularly in open team-building formats like OU where opponent Pokémon attributes are initially unknown. The bot maintains a comprehensive dataset of possible sets for each species, sourced from Smogon usage statistics, scraped team builds from forums, and public replay data. As battles progress, Foul Play intelligently infers hidden information through game mechanics: damage calculations reveal stat distributions, move priority ordering constrains speed ranges, extended weather duration indicates specific items, and move usage patterns eliminate equipment possibilities (e.g., status move usage eliminates Assault Vest; hazard damage reactions exclude Heavy-Duty Boots). By sampling from likelihood-weighted distributions of possible opponent configurations during search, the bot adapts its strategy based on the most probable scenarios.

This combination of efficient search and accurate prediction enabled Foul Play to achieve strong results: over 90\% GXE in Generation 9 Random Battles (peak rating 2341), over 80\% GXE in Generation 9 OU (peak rating 1879), and first place in the Pok\'eAgent Challenge Gen 9 OU track with a dominant 50--14 finals victory.

\begin{figure}[h]
    \centering
    \includegraphics[width=0.8\textwidth]{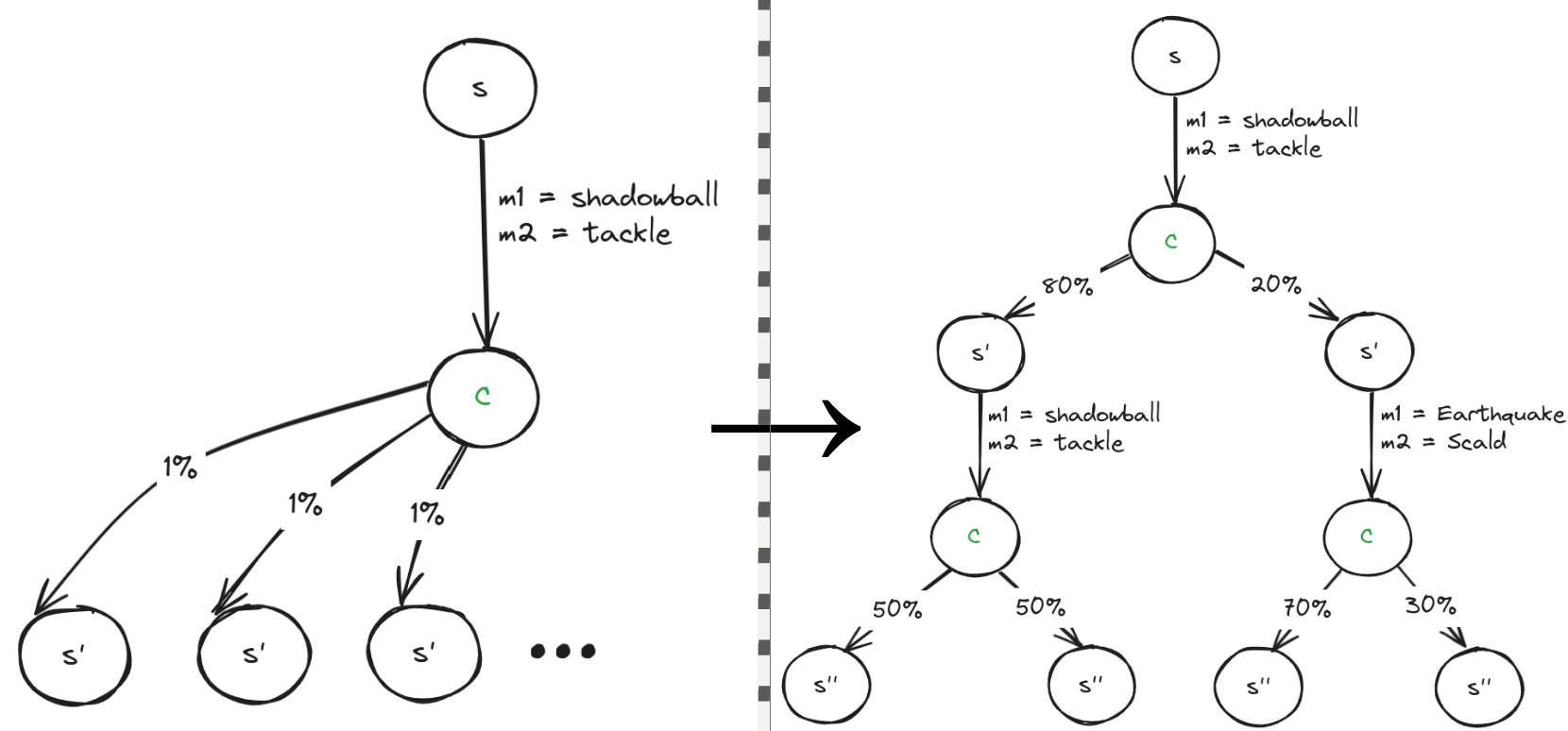}
    \caption{Foul Play's MCTS search architecture with damage roll grouping. The engine clusters damage outcomes by knockout potential rather than exploring all 32 possible values (16 rolls $\times$ critical hit), enabling deeper search while preserving strategic relevance.}
    \label{fig:foulplay-search}
\end{figure}

\begin{figure}[h]
    \centering
    \includegraphics[width=0.6\textwidth]{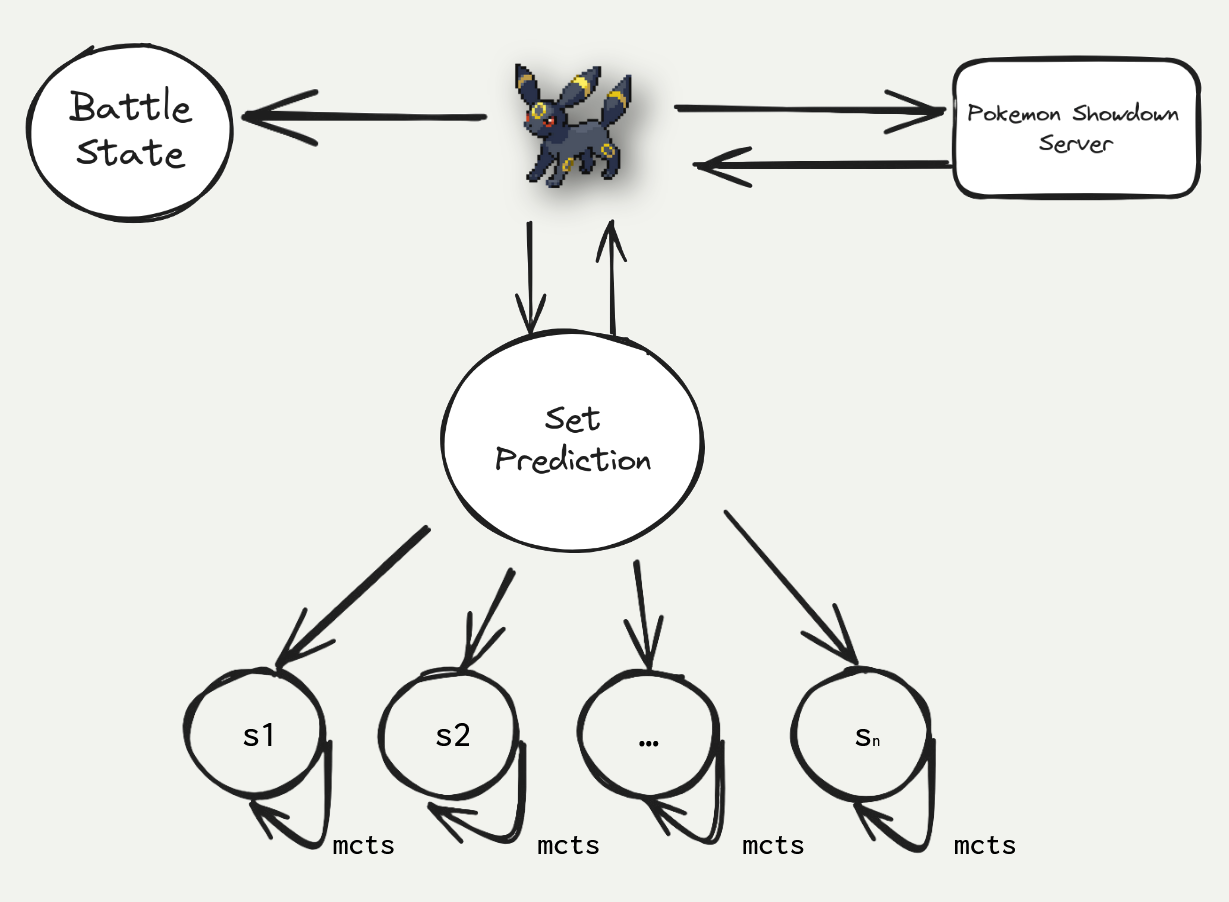}
    \caption{Foul Play's set prediction pipeline. Hidden information is progressively inferred through damage calculations, priority ordering, weather duration, and move usage patterns, narrowing the space of possible opponent configurations.}
    \label{fig:foulplay-prediction}
\end{figure}

\subsubsection{4thLesson (Gen 1 OU Finalist)}
\label{sec:4thlesson}

\textbf{Team:} Gyungbo Kim, Sangyeon Park, Eunju Kwon, Yujin Kim \\
\textbf{Affiliation:} Pok\'eAgent Challenge Gen 1 OU Finalist

4thLesson follows Metamon's SyntheticRL-V2 model and adopts a fine-tuning approach starting from its pretrained model, while preserving the original architecture for stable initialization. Two key modifications distinguish their approach:

First, they employ the Kron (Kronecker-factored Approximate Curvature) optimizer instead of the AdamW optimizer originally used in the Amago framework. Although the Kron optimizer is a second-order optimizer and incurs higher computational cost than AdamW, recent work demonstrates that it stabilizes reinforcement learning by ensuring more consistent gradient flow when scaling up model size, outperforming Adam-based optimizers in this regard.

Second, they adopt AID (Activation by Interval-wise Dropout) instead of the previously used Leaky ReLU. AID introduces additional linearity into the model, mitigating plasticity loss in continual learning and reinforcement learning settings where model plasticity tends to degrade. Since AID also acts as a form of dropout, they removed the dropout modules used in the original model.

To collect high-quality self-play data, they employ a multi-stage data generation strategy based on a local ladder setup. They first generated self-play replays using the 19 baseline models provided by a local ladder setup, collecting approximately 30k replays for small-size models, 40k for medium-size models, and 50k--100k for large-size models. All replays were generated using the modern\_replays (v1) teamset, resulting in roughly 800k replays in total. In addition, they performed self-play between intermediate checkpoints of their own models, collecting another 300k--400k replays and increasing the total dataset size to approximately 1.1--1.2M replays. After the preliminary round, they repeated a similar process, generating about 10k samples per model, using the modern replays (v2) teamset and adding approximately 130k more samples, increasing the total dataset size to 1.3--1.4M samples.

\begin{figure}[h]
    \centering
    \includegraphics[width=0.9\textwidth]{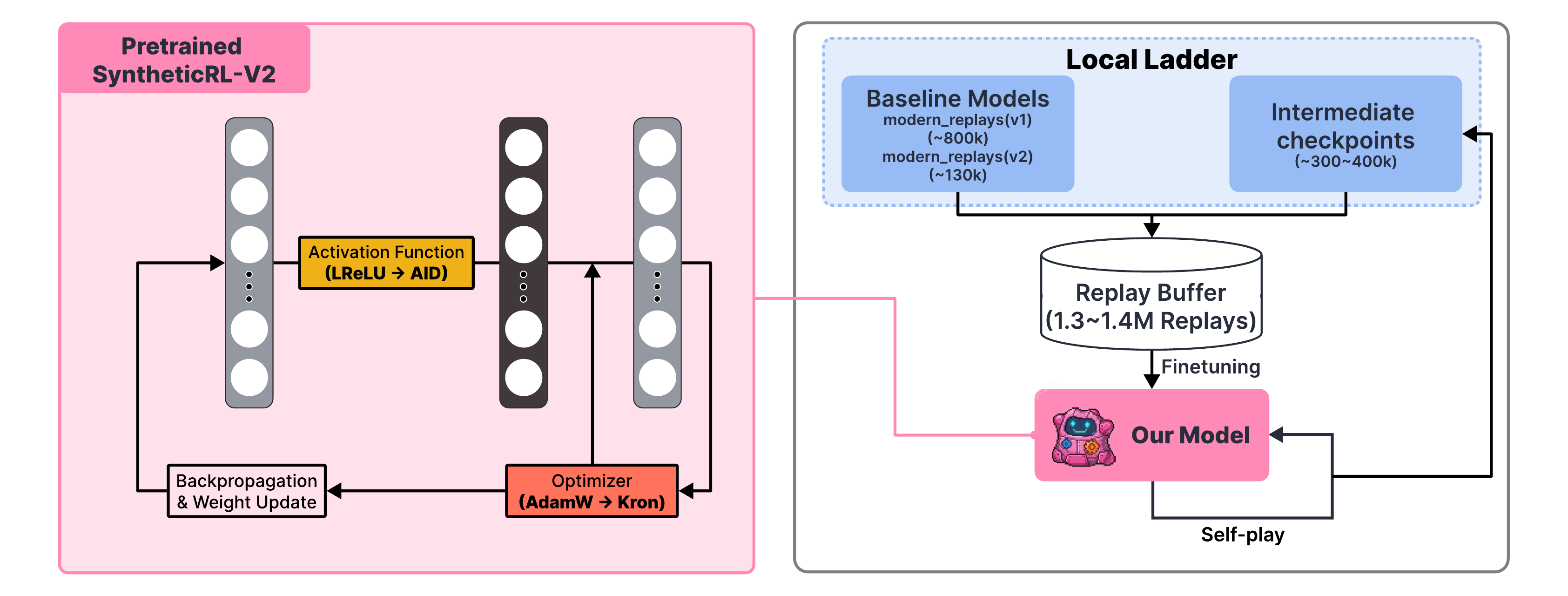}
    \caption{4thLesson's learning architecture and training pipeline, showing the integration of the Kron optimizer and AID activation function.}
    \label{fig:4thlesson-method}
\end{figure}

\begin{figure}[h]
    \centering
    \includegraphics[width=0.6\textwidth]{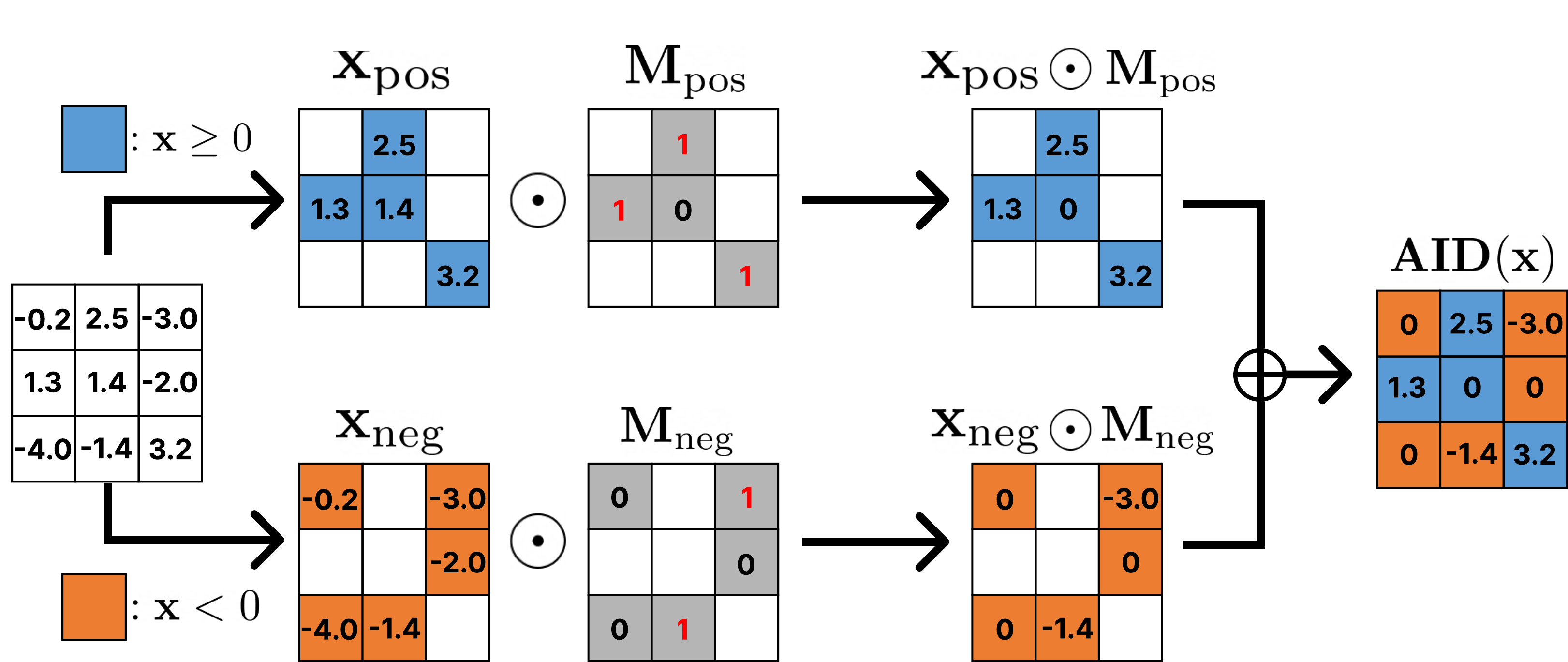}
    \caption{Structure of Activation by Interval-wise Dropout (AID) used by 4thLesson to mitigate plasticity loss.}
    \label{fig:4thlesson-aid}
\end{figure}

\subsubsection{Team Q (Gen 9 OU Finalist)}
\label{sec:team-q}

\textbf{Team:} Qiao Wang, Ling Wu \\
\textbf{Affiliation:} Pok\'eAgent Challenge Gen 9 OU Finalist

Team Q's approach centers on a Two-Phase Curriculum Learning framework designed to bootstrap basic competency before refining advanced strategic reasoning. The agent architecture utilizes a 50M parameter Actor-Critic model. The core innovation lies in splitting the training process into two distinct phases: a \textit{Mechanics Phase} and a \textit{Strategy Phase}.

In the initial Mechanics Phase, the agent is fine-tuned against a suite of heuristic bots. This phase focuses on learning the fundamental rules of the game, such as type advantages, move effectiveness, and valid action masking, without the noise of complex adversarial behavior. Once the agent achieves a baseline win rate against these static policies, it transitions to the Strategy Phase. Here, they employ an iterative ``coach cycle'' where the agent trains against a fixed expert coach and previous versions of itself. This self-play and expert-play hybrid allows the agent to develop deeper foresight and prediction capabilities, resulting in a steady increase in win rates from version v0 to v6 as observed in their experiments.

Their results demonstrate that the separation of mechanics learning from strategic fine-tuning significantly accelerates convergence. The agent showed the most dramatic performance jumps when transitioning from the v0 team set (random/heuristic initialization) to the v1 team set (curriculum-based), validating the hypothesis that mastering low-level game dynamics is a prerequisite for high-level long-horizon planning in stochastic environments like Pokémon.

\begin{figure}[h]
    \centering
    \includegraphics[width=0.65\textwidth]{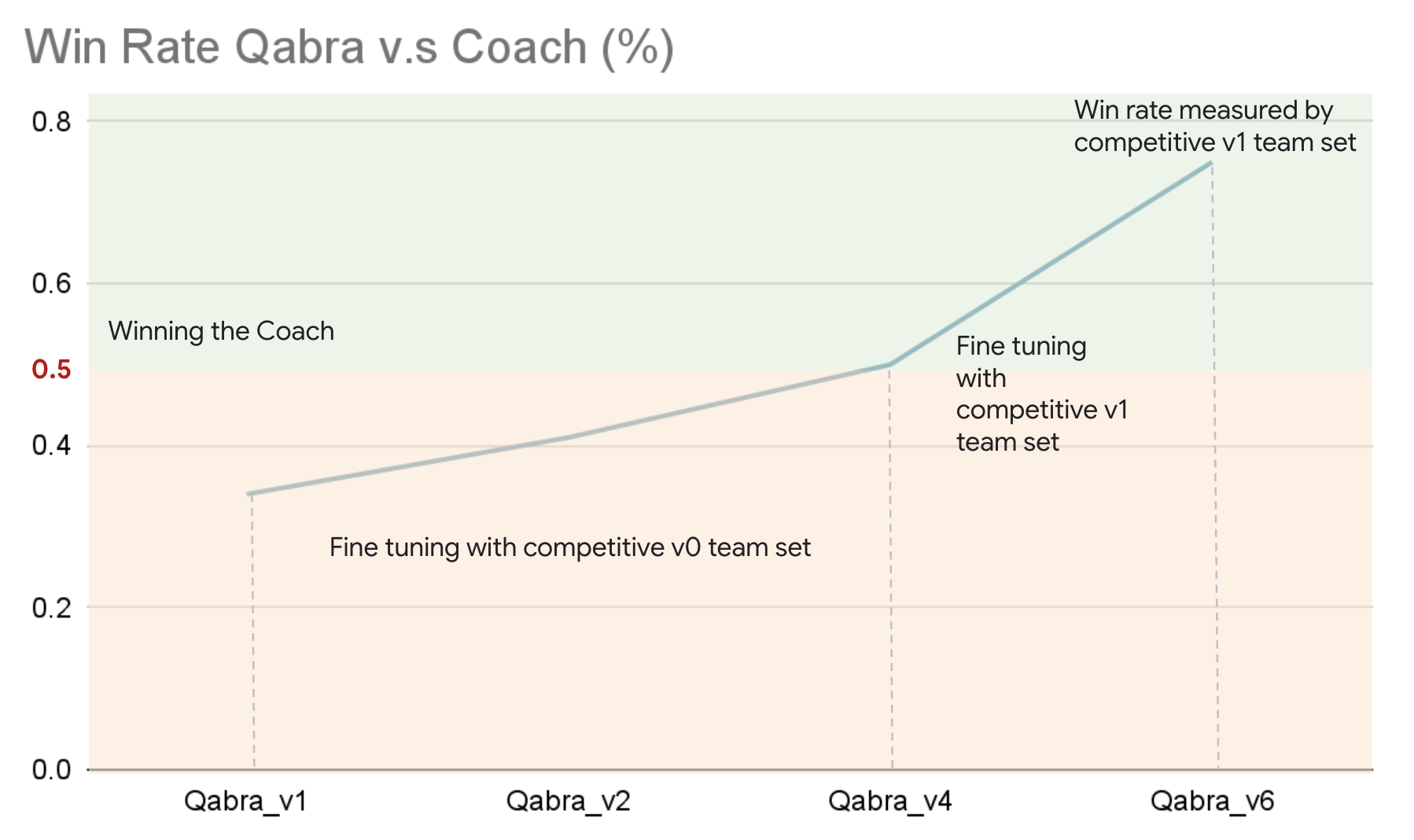}
    \caption{Team Q's win rate performance across training iterations (v1--v6). Performance is measured against the Competitive v1 Team Set. The initial iterations (v1--v4) were trained on the v0 team set, while v6 incorporates fine-tuning on the v1 set, resulting in a performance jump to approximately 0.8.}
    \label{fig:team-q-winrate}
\end{figure}

\subsection{Speedrunning Track: RPG Speedrunning}

\subsubsection{Heatz (Speedrunning Track Champion)}
\label{sec:heatz}

\textbf{Author:} Junik Bae \\
\textbf{Affiliation:} Pok\'eAgent Challenge Speedrunning Track Champion

\begin{figure}[h]
  \centering
  \includegraphics[width=\textwidth]{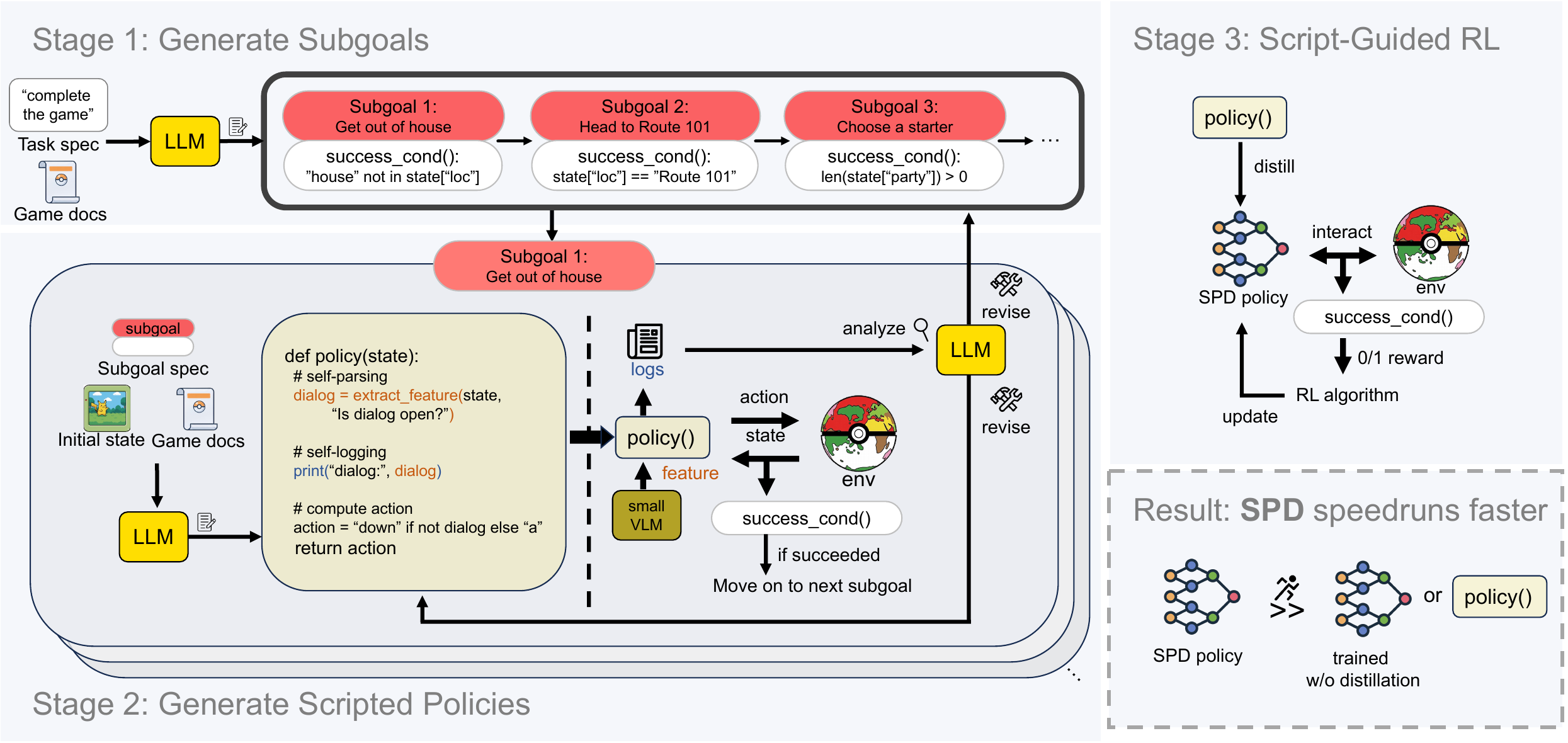}
  \caption{\textbf{Overview of Scripted Policy Distillation (SPD).} The approach consists of three stages: (1) Subgoal Generation where an LLM decomposes the task into sequential subgoals with executable success conditions; (2) Scripted Policy Generation where the LLM generates policies that can invoke VLM tools and use self-directed logging for debugging; (3) Script-Guided RL where scripted policies are distilled into neural networks via supervised learning followed by RL with expert action guidance.}
  \label{fig:heatz-method}
\end{figure}

Reinforcement learning (RL) offers a powerful framework for training autonomous agents, yet applying RL directly to long-horizon tasks with sparse rewards remains fundamentally challenging. To address this challenge, we propose leveraging LLM-generated scripted policies as priors for RL exploration (Figure~\ref{fig:heatz-method}). Although these scripted policies are not always optimal, they facilitate RL training by initializing exploration from a distribution that already reaches the goal, rather than from scratch. We implement this long-horizon task learning framework as \textbf{Scripted Policy Distillation (SPD)}, which consists of three stages: (1) subgoal generation~\citep{brohan2023can, zhu2023ghost}, (2) scripted policy generation~\citep{liang2023code, wangvoyager}, and (3) script-guided RL.

\paragraph{(1) Subgoal generation.} Given a long-horizon task specification, an LLM decomposes the task into a sequence of subgoals, each paired with an executable success-condition function \texttt{success\_cond(state)} that determines subgoal completion.

\paragraph{(2) Scripted policy generation.} For each subgoal, the LLM generates a \textit{scripted policy} that maps states to actions. The scripted policy interacts with the environment until \texttt{success\_cond(state)} returns \texttt{True} or a timeout occurs. Upon failure, the LLM analyzes execution traces and revises either the policy code (Stage 2) or the subgoal specification (Stage 1).

This stage employs two key techniques to handle complex environments. First, to extract rich visual cues that are not present in the state inputs, policies can optionally query a vision-language model (Qwen3-VL-8B) and use the resulting information for decision making. Second, to efficiently update subgoals and scripted policies, we provide the LLM with concise summaries of agent interactions obtained via \textit{self-directed logging}, rather than long, full execution trajectories.

\paragraph{(3) Script-guided RL.} Once all scripted policies reliably achieve their corresponding subgoals, we distill them into neural network policies via imitation learning on expert trajectories, followed by RL with expert action guidance. The resulting neural policies execute faster than the scripted policies and can therefore discover more efficient strategies.

For distillation, we train a DQN agent with two forms of expert guidance. First, we seed the replay buffer with successful expert trajectories. Second, during rollouts, we execute expert actions with probability $\epsilon$ (annealed from $0.1$ to $0$), while otherwise following the learned policy. Together, these mechanisms bootstrap learning from expert behavior while allowing improvements through RL optimization.

Our method, SPD, achieved a 40:13 run on Pokémon Emerald up to the first gym. The resulting policy exhibited several interesting emergent behaviors across the pipeline. During scripted policy generation, the agent occasionally synthesized explicit search routines, such as BFS over a local navigation graph, to reliably plan short paths for navigation subgoals. Furthermore, after distillation and RL fine-tuning, the neural policy executed these behaviors more efficiently and further improved speed through strategies not explicitly programmed in the scripts, including shorter routes, skipping unnecessary trainer battles, and faster menu interactions.

\subsubsection{Hamburg Pok\'eRunners (Speedrunning Track Second Place)}
\label{sec:hamburg}

\textbf{Team:} Benedikt Schink, Arian Urdu, Matin Urdu \\
\textbf{Affiliation:} Pok\'eAgent Challenge Speedrunning Track Second Place

Hamburg Pok\'eRunners achieved second place (first badge: 01:14:43) using a reinforcement learning approach based on recurrent proximal policy optimization (PPO). They used a standard recurrent PPO architecture with a long short-term memory (LSTM) in the recurrent state. The encoder consists of a convolutional neural network (CNN) and a multi-layer perceptron (MLP). The CNN receives downsampled game frames (grayscale, $128 \times 128$ pixels). A sinusoidal positional encoding is used to encode the $x$ and $y$ in-game coordinates. The location (e.g., the current city or route) is encoded using a one-hot encoding. Together, the location and the sinusoidal positional encoding result in unique positional coordinates. This global conditioning prevents the spatial ambiguity inherent in local coordinate systems.

The last input of the MLP is a binary milestone vector $m \in \{0, 1\}^{38}$, which encodes the sub-milestones of the competition, as well as the locations and Pok\'ecenters reached along the way. The milestone vector serves both as a memory component, keeping track of the milestones achieved thus far, and as goal-conditioning, keeping track of the current objective. This played a crucial role in making the model more stable to the inherent randomness of Pokémon Emerald and the non-deterministic input/output latency of the evaluation script. They observed simple error-correcting behavior, where the agent would overshoot or undershoot the target, resulting in backtracking to correct the trajectory.

Their reward function includes only positive rewards that incentivize desired behavior: competition sub-milestones, locations, coordinate exploration, badges received, money earned, HP healed, Pokémon leveled up, and damaging attacks used. All negative rewards they tried, such as step penalties, resulted in model degradation as the model reverted to only safe behavior and did not progress in the game.

Due to limited computational resources, they trained multiple smaller models that were then stitched together during inference, instead of training one larger model. Most attempts at transfer learning yielded only limited success, mostly helping with very basic navigation capabilities. While transfer learning yielded only small improvements restricted to basic navigation, the inclusion of a milestone vector significantly mitigated catastrophic forgetting.

\begin{figure}[h]
    \centering
    \includegraphics[width=0.8\textwidth]{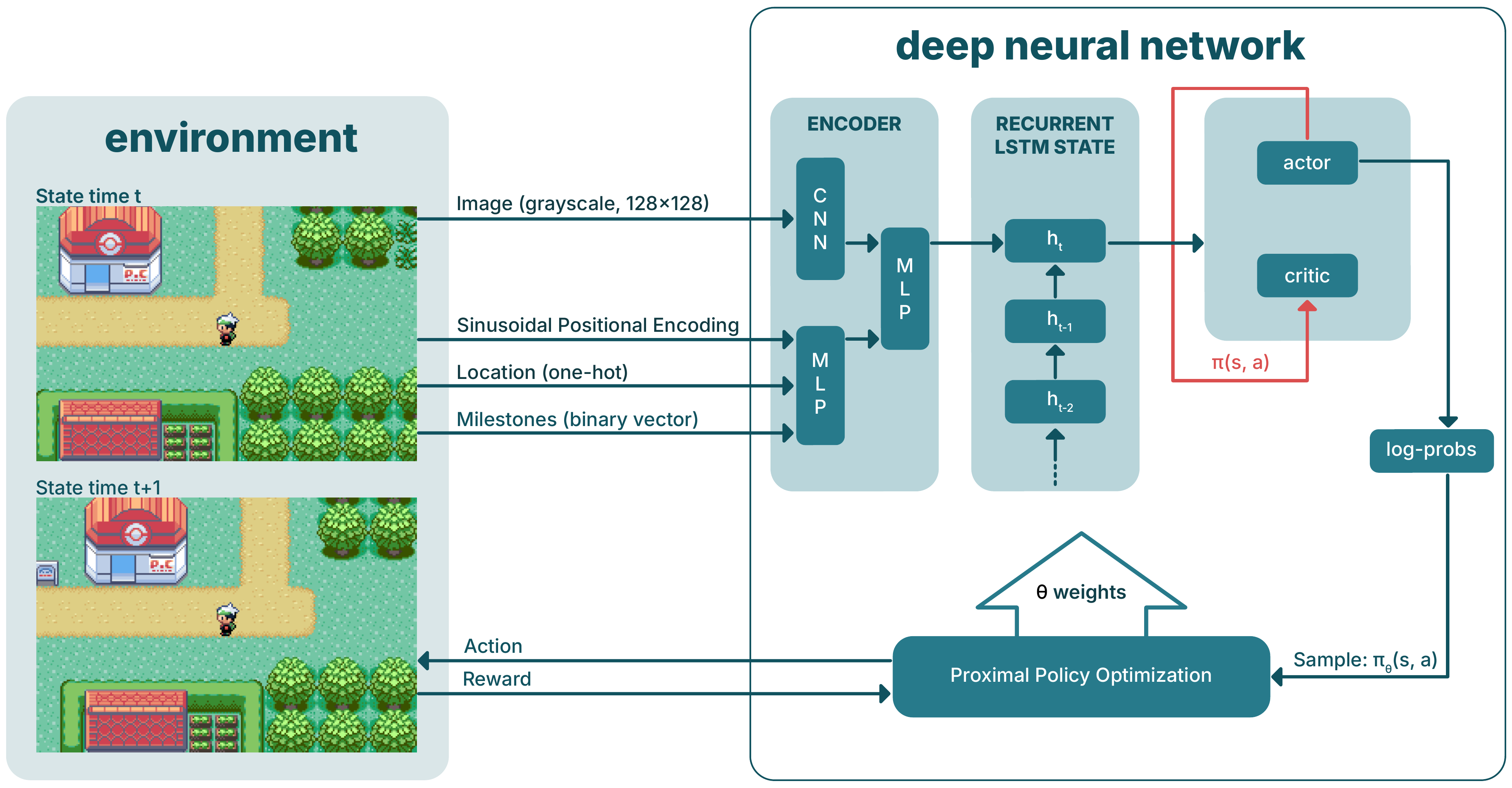}
    \caption{Hamburg Pok\'eRunners' recurrent PPO architecture, showing the CNN encoder for visual input, MLP for positional and milestone encoding, and LSTM for temporal reasoning.}
    \label{fig:hamburg-architecture}
\end{figure}

\subsubsection{Anthonys (Speedrunning Track Third Place)}
\label{sec:anthonys}

\textbf{Author:} Anthony Sistilli \\
\textbf{Affiliation:} Pok\'eAgent Challenge Speedrunning Track Third Place

Anthonys' approach centered on decomposing the game progression into discrete phases with specialized navigation strategies for each context. The core innovation was combining deterministic pathfinding algorithms with context-aware prompt engineering to guide the language model through complex navigation and battle scenarios.

\textbf{Phase-Based State Management:} They structured the challenge into seven distinct phases, each corresponding to major game milestones. Each phase maintained its own prompt template with conditional instructions that dynamically adapted based on completed objectives and current location.

\textbf{A* Pathfinding with Directional Priorities:} Rather than relying solely on the LLM for spatial reasoning, they implemented A* pathfinding on the game map grid to compute optimal movement sequences. The system extracted walkable tiles from game memory, constructed a navigable grid representation, and generated action sequences to reach target coordinates.

\textbf{Battle State Management:} Battle sequences required special handling due to the game's menu system complexity. They implemented an input-clearing mechanism that sent predetermined button sequences to ensure the agent escaped bad states.

\textbf{Adaptive Recovery and Stuck Detection:} The system tracked position history to detect when the agent remained stationary for multiple turns, indicating an obstacle or navigation failure.

\subsubsection{Deepest (Speedrunning Track Judge's Choice: Most Efficient)}
\label{sec:deepest}

\textbf{Team:} Seonghun Hong, Hyunyoung Jeong, Hyeokgi Kim, Jaeyoon Jung \\
\textbf{Affiliation:} Seoul National University, Soongsil University, MAUM.AI

\begin{figure}[h]
    \centering
    \includegraphics[width=0.92\textwidth]{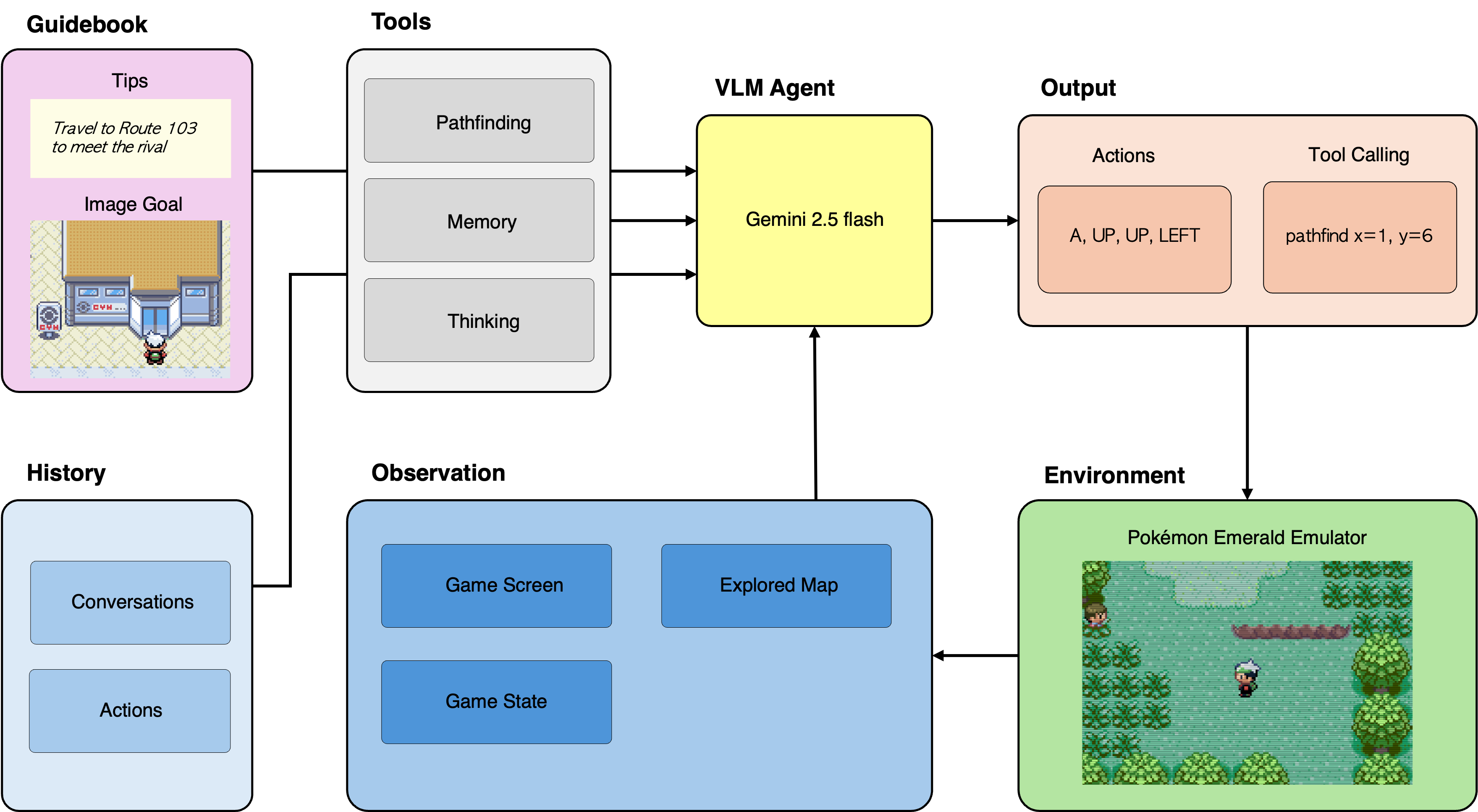}
    \caption{Deepest agent architecture overview. The agent receives partial observations and milestone-specific guidance from the Guidebook. Gemini 2.5 Flash reasons and outputs actions, optionally invoking tools (Pathfinding, Memory, Thinking). Actions execute in the emulator, producing the next observation. History provides temporal context.}
    \label{fig:deepest-architecture}
\end{figure}

We propose a training-free, fully autonomous agent architecture for Pokémon Emerald speedrunning that operates under human-like perceptual constraints using the Gemini 2.5 Flash vision-language model. Our approach achieved 5th place with a completion time of 02:04:29 for the first gym milestone and received the \textbf{Judge's Award} for most sample-efficient completion, demonstrating the effectiveness of principled agent design without data collection or human feedback.

A core design principle of our system is the absence of privileged game state. The agent receives no ground-truth map information; all navigation decisions---including those made by our pathfinding module---are computed exclusively from partially observed tiles explored during gameplay. This constraint closely mirrors human visual perception and ensures that planning and control emerge from observation-driven reasoning rather than access to internal game mechanics. To support high-level strategic decision-making, we introduce a \textbf{Guidebook} system inspired by how human speedrunners consult established guides. The guidebook provides milestone-specific knowledge, such as traveling to Route 103 to meet the rival and selecting Mudkip for type advantage against the Rock-type gym leader. For goal specification, we employ \textbf{image goals}---reference images of destination locations such as Professor Birch's Lab and Dad's Gym, which are visually indistinct buildings that are difficult to identify without prior game knowledge. These image goals enable the agent to visually recognize objectives and ground its actions in perceptual similarity. The overall architecture is illustrated in Figure~\ref{fig:deepest-architecture}.

Rather than relying solely on low-level action tokens (e.g., ``UP'', ``LEFT'', ``A''), we augment the agent with a set of auxiliary tools that allow it to autonomously reason, plan over long horizons, and adapt its computation, thereby better leveraging its decision-making capacity for efficient speedrunning. Specifically, we introduce three auxiliary tools:

\textbf{Pathfinding Tool:} This tool provides two navigation interfaces: coordinate-based navigation for known destinations, and directional exploration when exact coordinates are unknown. The pathfinder computes optimal trajectories on the observed map grid with the A* algorithm, incorporating game-specific constraints including one-way ledges and dynamic NPC collision avoidance.

\textbf{Memory Tool:} This tool supports long-horizon planning by enabling the agent to explicitly persist critical state information. To prevent the accumulation of stale or task-irrelevant information, we apply a sliding window policy that retains memory only over the most recent 10 conversation turns.

\textbf{Thinking Tool:} This tool implements adaptive computation by toggling the model's reasoning budget---allocating 2048 tokens for complex decisions while defaulting to 512 tokens for routine navigation.

\section{Environment and System Details}
\label{sec:environment-details}

\subsection{RPG System Comparison}
\label{sec:system-comparison}

We analyze Pokémon RPG AI systems through a five-dimensional categorical framework $\mathcal{S}(\mathcal{A}) = (S, T, M, F, \Phi)$, characterizing state representation, tools, memory architecture, feedback mechanisms, and fine-tuning approach. Table~\ref{tab:system-comparison} illustrates this decomposition for each system as it was initially popularized.\footnote{These systems are under active development and have since adopted features---minimal additional information and self-reflection mechanisms---that were initially introduced by the \pokeagent project. We characterize each system at the version contemporaneous with PokeAgent's development (mid 2025), as later iterations increasingly converge on similar design choices (such as using sub-agents for feedback) without being completely open-source at the time of writing.} For current version capability for X Plays Pokémon, we encourage readers to directly view the stream of the individual creators. Many of these X Plays Pokémon baselines have completed numerous distinct Pokémon RPG games, which is very impressive from the view of the ML community. We recommend standardization among subsections of the game for easier comparison with state standardized by PokéAgent. For X Plays Pokémon, systems vary dramatically across each dimension: state representations range from visual frames to structured RAM extraction; tool availability ranges from basic button inputs to full planning suites with code execution; memory architectures span raw context windows to engineered summarization pipelines. Critically, even the reported evaluation metrics---variously ``steps,'' ``actions,'' and ``turns''---lack shared definitions across projects. These design choices dramatically affect measured performance, making cross-system comparison scientifically problematic without standardization.

\begin{table}[h]
\centering
\caption{Analysis of Pokémon RPG AI systems using the $(S, T, M, F, \Phi)$ framework, characterized at the version each system was initially popularized. Heterogeneity across dimensions makes direct performance comparison methodologically unsound without standardization.}
\label{tab:system-comparison}
\resizebox{\textwidth}{!}{
\begin{tabular}{@{}lccccc@{}}
\toprule
\textbf{System} & \textbf{State $(S)$} & \textbf{Tools $(T)$} & \textbf{Memory $(M)$} & \textbf{Feedback $(F)$} & \textbf{Fine-tuning $(\Phi)$} \\
\midrule
Claude Plays Pokémon\textsuperscript{$\dagger$} & Frame + location, walkable tiles, party stats & Pathfinder, knowledge base & File-based summaries + knowledge base & ReAct loop & Zero-shot \\
GPT Plays Pokémon\textsuperscript{$\dagger$} & Frame + location, party stats, money, tile colors & Button inputs, self-built minimap & Goal tracker + notepad & Varied & Zero-shot \\
Gemini Plays Pokémon\textsuperscript{$\dagger$} & Frame + object positions, tile properties, navigability & Self-generated tools, code exec, pathfinder agent & Notepad + map markers + mental map & Varied & Zero-shot \\
Nunu AI & Frame + full parsed game state (map, party, items, NPCs) & Navigation, planning tools & Persistent memory store & ReAct loop + Twitch chat & Zero-shot \\
CLI Agents\textsuperscript{$\ddagger$} & Frame + map, party, bag & Game MCP tools (get state, press buttons, pathfinding) & Agent context window & ReAct loop & Zero-shot \\
\textbf{PokeAgent} & Frame + map, party, bag & Full MCP tool suite (pathfinding, memory, progress summary, reflection, etc)& knowledge base +  Action history window& Multi-agent self-reflection & Zero-shot \\
\bottomrule
\multicolumn{6}{l}{\textsuperscript{$\dagger$}Community-built harnesses, not official products of the respective model providers.} \\
\multicolumn{6}{l}{\textsuperscript{$\ddagger$}Claude Code, Codex CLI, Gemini CLI --- evaluated on the \pokeagent Speedrunning Track with standardized MCP tool access.} \\
\end{tabular}
}
\end{table}

The key architectural distinction is in feedback and reflection. The ``X Plays Pokémon'' systems operate as observe-act loops: the model receives an observation, selects an action, and observes the result, with no explicit mechanism for evaluating its own decisions or recovering from errors. In contrast, \pokeagent employs a multi-agent self-reflection system in which a dedicated critic agent evaluates action outcomes, detects suboptimal play, and triggers strategy revision---enabling error recovery rather than error compounding. This distinction is not merely taxonomic: pure observe-act agents are known to exhibit ``panic behavior''~\citep{gemini2p5report}, where a single tactical mistake cascades into increasingly poor decisions, a failure mode that self-reflection explicitly mitigates.

A further distinction is in execution model. The ``X Plays Pokémon'' systems all freeze the emulator between actions, giving the model unlimited deliberation time per step---wall-clock runtimes of hundreds of hours are common (e.g., Gemini's 406-hour playthrough of Pokémon Blue). None penalize slow inference in their reported metrics, fully decoupling computational cost from any notion of real-time play. The \pokeagent Speedrunning Track, by contrast, measures wall-clock time as a primary metric: the emulator runs continuously and agents must issue actions in real time, penalizing slow inference and long reasoning chains. This makes speedrunning performance a joint measure of decision quality \emph{and} computational efficiency, more closely reflecting the practical constraints of deploying agents in interactive environments.

Among contemporary RPG systems, progress has been rapid but fragmented.
By early 2026, frontier models had completed full Pokémon playthroughs across multiple games---including Red, Blue, Crystal, and Emerald---using a variety of harness architectures.
Notable milestones include Gemini 2.5 Pro completing Pokémon Blue in approximately 406 hours~\citep{gemini2p5report,zhang2025geminiplayspokemon}, GPT-5 finishing Pokémon Red in 6,470 steps~\citep{engadget2025gptplayspokemon}, Gemini 3 Pro completing Pokémon Crystal~\citep{zhang2025gemini3playspokemon}, and later model iterations (GPT-5.1, GPT-5.2, Gemini 3 Flash) completing Crystal and Emerald as well.
Claude Plays Pokémon~\citep{anthropic2025visible} used chain-of-thought reasoning with a harness to complete a small section of the game over 35,000 actions, where its predecessor without these capabilities failed to exit the starting location.
However, these achievements resist direct comparison: the underlying harnesses differ fundamentally in state representation, tool access, and action granularity, and even the reported metrics---variously ``steps,'' ``actions,'' and ``turns''---lack shared definitions across projects.
The \pokeagent Challenge Speedrunning Track addresses precisely this gap, providing a standardized state and evaluation protocol that enables verified, apples-to-apples comparisons across models and harnesses.

See Appendix~\ref{sec:baseline-architectures} for detailed architecture descriptions of our baselines, including the \pokeagent multi-agent orchestration system.

\subsubsection{Speedrunning: Evaluation Interface}

\begin{figure}[H]
    \centering
    \includegraphics[width=\linewidth]{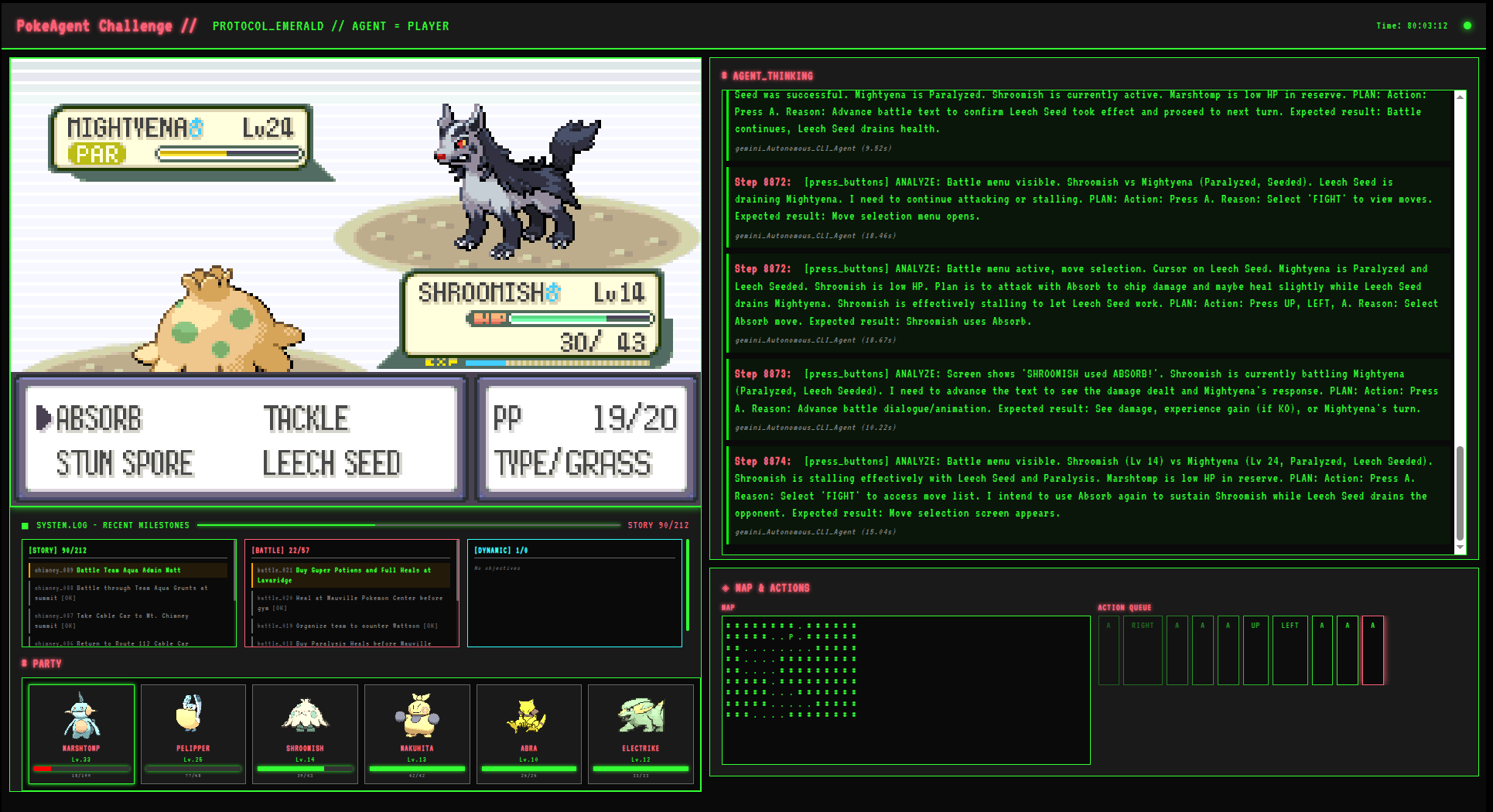}
    \caption{\textbf{Speedrunning Track Evaluation Interface.} The interface displays: (1) the Pokémon Emerald game screen showing a trainer battle with move selection (top left), (2) agent reasoning traces with timestamped decision logs detailing battle strategy and action rationale (right panel), (3) current party composition with sprite previews (bottom left), and (4) map overview with action history (bottom right). This visualization enables real-time debugging of agent behavior across the thousands of timesteps required for speedrunning.}
    \label{fig:track2_interface}
\end{figure}

\subsection{LLM Baseline Token Usage and Cost}
\label{sec:llm-cost}

Table~\ref{tab:llm-cost} reports token consumption and API cost for each LLM baseline on the Gen~9 OU Extended Timer ladder. Costs vary by over $70\times$ across models: GPT-5.2 is the most expensive at \$1.247 per game, while DeepSeek~V3 and Qwen3.5 Plus cost roughly \$0.015. Output token counts per turn also vary dramatically---MiniMax produces $\sim$6.2K completion tokens per turn, whereas DeepSeek~V3 and Qwen3.5 Plus produce fewer than 10. These differences highlight that raw per-game cost is driven primarily by output pricing and reasoning verbosity rather than game length.

\begin{table}[h]
\centering
\small
\caption{LLM baseline token usage and API cost on the Gen~9 OU Extended Timer ladder. Input/output tokens are per-turn averages. Models sorted by GXE rank (descending).}
\label{tab:llm-cost}
\resizebox{\textwidth}{!}{
\begin{tabular}{@{}lrrrrrr@{}}
\toprule
\textbf{Model} & \textbf{Avg Input Tok/Turn} & \textbf{Avg Output Tok/Turn} & \textbf{Avg Turns} & \textbf{Avg Cost/Game} & \textbf{Avg Cost/Turn} & \textbf{Total Cost} \\
\midrule
Gemini 3.1 Pro    & 1,245 &     54 & 21.1 & \$0.066  & \$0.0031 &   \$8.86 \\
GPT-5.2           & 2,472 &  4,270 & 17.8 & \$1.247  & \$0.0702 & \$162.16 \\
Gemini 3 Flash    & 1,265 &     76 & 23.7 & \$0.020  & \$0.0009 &   \$0.90 \\
GLM-5             & 1,658 &    960 & 18.3 & \$0.069  & \$0.0038 &   \$7.98 \\
Gemini 3 Pro      & 1,211 &    115 & 23.9 & \$0.091  & \$0.0038 &   \$3.72 \\
Claude Opus 4.6   & 2,110 &    137 & 24.4 & \$0.341  & \$0.0140 &  \$73.59 \\
Grok-3 Mini       & 1,902 &  1,262 & 23.7 & \$0.029  & \$0.0012 &   \$4.27 \\
Claude Sonnet 4.6 & 2,234 &    132 & 23.7 & \$0.206  & \$0.0087 &  \$34.83 \\
Grok-3            & 1,971 &      9 & 22.2 & \$0.134  & \$0.0061 &  \$24.96 \\
MiniMax M2.5      & 2,327 &  6,218 & 18.3 & \$0.120  & \$0.0065 &   \$1.31 \\
Hermes 4 405B     & 2,310 &     91 & 21.7 & \$0.056  & \$0.0026 &   \$7.22 \\
DeepSeek V3       & 2,238 &      9 & 23.5 & \$0.017  & \$0.0007 &   \$3.42 \\
Kimi K2.5         & 1,900 &  4,510 & 19.2 & \$0.207  & \$0.0108 &   \$3.72 \\
Qwen3.5 Plus      & 2,306 &      8 & 22.9 & \$0.014  & \$0.0006 &   \$2.55 \\
\bottomrule
\end{tabular}
}
\end{table}

\subsection{Speedrunning Track Task Diversity}

\begin{figure}[h]
    \centering
    \includegraphics[width=\linewidth]{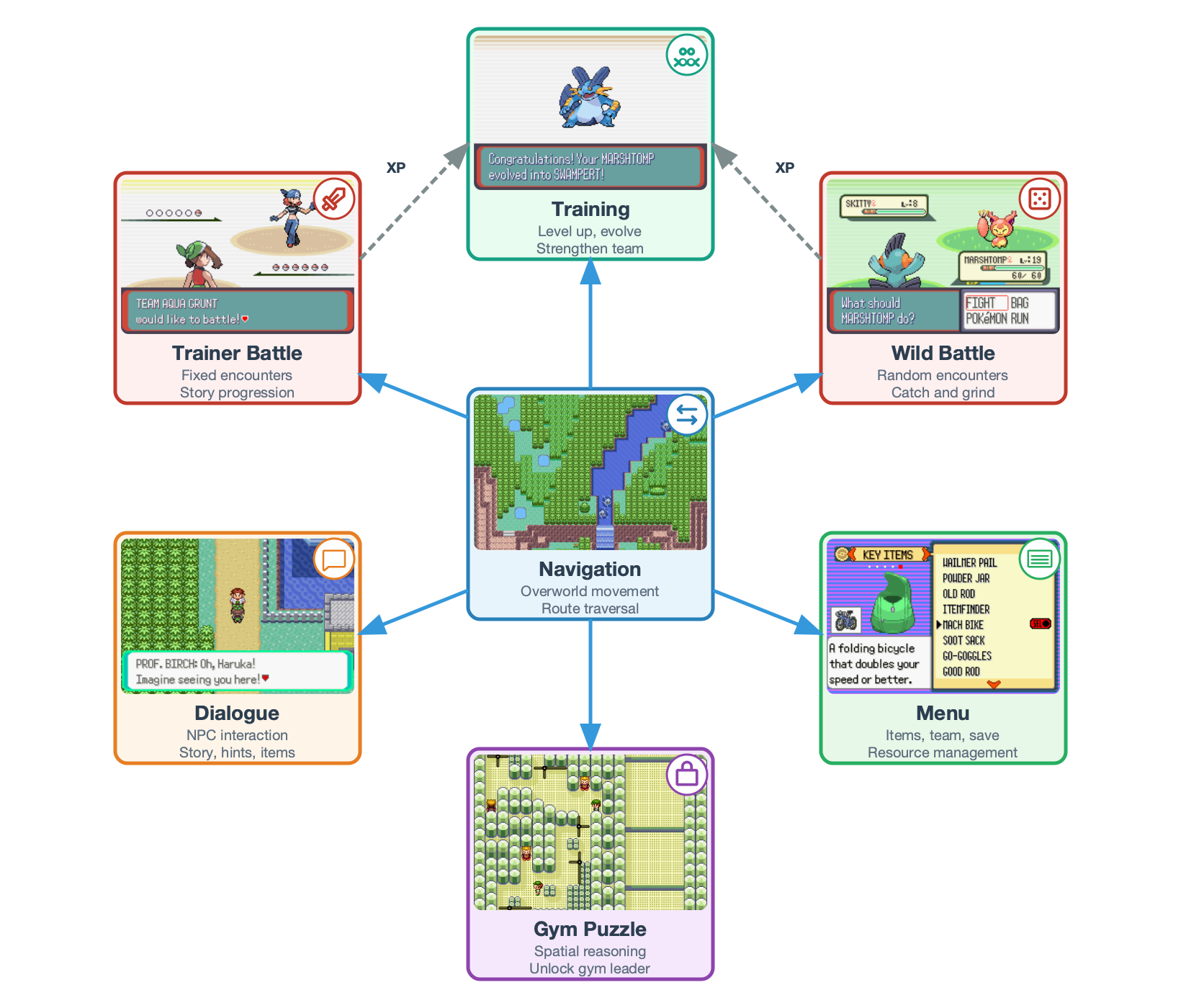}
    \caption{\textbf{Speedrunning Track Task Diversity.} Directed graph showing RPG subtask categories and their dependencies. Core tasks include overworld navigation, battle encounters (wild and trainer), gym puzzles, menu management, and NPC dialogue. Edges indicate how completing one task type enables or requires others---for example, navigation leads to encounters, battles yield experience for team building, and menu interactions manage resources needed for all other tasks.}
    \label{fig:track2_tasks}
\end{figure}

\section{State Space Complexity Derivation}
\label{sec:state-space-derivation}

This appendix derives combinatorial upper bounds on Pokémon team configuration and battle state spaces for Gen 1 OU, Gen 9 OU, and Gen 9 VGC.
We clearly distinguish \emph{exact} quantities (marked \textbf{E}), \emph{exact upper bounds} (marked \textbf{E upper bound}), and \emph{approximations} (marked \textbf{A}), and justify every choice.
Moves unavailable in Generation~9 (\texttt{isNonstandard:\ "Past"} in the Pok\'emon Showdown source) are excluded from the Gen~9 derivations.
All arithmetic has been verified programmatically.

\subsection{EV Spread Counting}
\label{sec:ev-derivation}

\textbf{Setup.}
Each Pokémon distributes Effort Values (EVs) across 6 stats.
EVs only affect the stat formula in multiples of 4, so we count in units of 4.
Let $x_i \in \mathbb{Z}_{\geq 0}$ be the number of 4-EV units allocated to stat $i$.
The constraints are:
\begin{equation}
    \sum_{i=1}^{6} x_i \leq 127, \qquad 0 \leq x_i \leq 63.
    \label{eq:ev-constraint}
\end{equation}
The upper bounds reflect the per-stat cap of $252\,\text{EV} = 63 \times 4$ and the total budget of $508\,\text{EV} = 127 \times 4$.

\textbf{Exact count via inclusion-exclusion.}
Introduce slack variable $x_7 \geq 0$ so that $\sum_{i=1}^{7} x_i = 127$.
Stars-and-bars without the per-stat cap gives $\binom{133}{6} = 6{,}856{,}577{,}728$.
Applying inclusion-exclusion on $x_i \geq 64$ (substitute $y_i = x_i - 64$, remaining sum $= 63$) gives $\binom{69}{6} = 119{,}877{,}472$ per capped stat.
Two simultaneously capped stats would require total $\geq 128 > 127$, so higher-order terms vanish.

\begin{equation}
    \boxed{|\mathcal{E}| = \binom{133}{6} - 6\binom{69}{6} = 6{,}856{,}577{,}728 - 719{,}264{,}832 = 6{,}137{,}312{,}896.}
    \label{eq:ev-exact}
\end{equation}
\textbf{(E)} All arithmetic exact; no approximation.

\subsection{Generation 9 Team Configuration Space}
\label{sec:gen9-derivation}

This bound applies to both Gen 9 OU and Gen 9 VGC, which draw from the same Pokémon pool.

\paragraph{Species \textbf{(E)}.}
\texttt{pokedex.ts} contains 1{,}329 standard entries for Generation 9: 1{,}025 base Pokémon (National Dex \#1--1025) plus 304 alternate formes.
Under Species Clause (no duplicates, unordered team of 6):
\begin{equation}
    \binom{1329}{6} = 7{,}566{,}741{,}017{,}135{,}560 \approx 7.57 \times 10^{15}.
\end{equation}

\paragraph{Movesets \textbf{(E upper bound)}.}
Mew has the largest movepool in Generation 9: exactly \textbf{375 moves}.
Using Mew's pool as the upper bound for all 6 team members:
\begin{equation}
    \binom{375}{4}^6 = 810{,}855{,}375^6 \approx 2.84 \times 10^{53}.
\end{equation}

\paragraph{Abilities \textbf{(E upper bound)}.}
Maximum 3 abilities per species (two standard plus one Hidden Ability):
\begin{equation}
    3^6 = 729.
\end{equation}

\paragraph{Individual Values \textbf{(E)}.}
Each stat has an IV in $\{0,\ldots,31\}$:
\begin{equation}
    \bigl(32^6\bigr)^6 = 32^{36} = 2^{180} \approx 1.53 \times 10^{54}.
\end{equation}

\paragraph{Effort Values \textbf{(E)}.}
Using the exact count from Section~\ref{sec:ev-derivation}:
\begin{equation}
    |\mathcal{E}|^6 = 6{,}137{,}312{,}896^6 \approx 5.34 \times 10^{58}.
\end{equation}

\paragraph{Natures \textbf{(E)}.}
25 natures exist, but 5 are neutral (all producing identical stat modifiers in battle), yielding $20 + 1 = 21$ functionally distinct natures:
\begin{equation}
    21^6 = 85{,}766{,}121.
\end{equation}

\paragraph{Held Items \textbf{(E)}.}
Generation 9 contains 248 standard held items:
\begin{equation}
    248^6 = 232{,}653{,}764{,}952{,}064 \approx 2.33 \times 10^{14}.
\end{equation}

\paragraph{Terastallization \textbf{(E)}.}
Each Pokémon may Terastallize into any of 19 types (the 18 standard types plus Stellar):
\begin{equation}
    19^6 = 47{,}045{,}881.
\end{equation}

\paragraph{Gen 9 team space.}
\begin{align}
    \mathcal{T}_{\text{Gen9}}
    &\approx \underbrace{7.57 \times 10^{15}}_{\text{species}} \times
       \underbrace{2.84 \times 10^{53}}_{\text{moves}} \times
       \underbrace{7.29 \times 10^{2}}_{\text{abilities}} \times
       \underbrace{1.53 \times 10^{54}}_{\text{IVs}} \times \notag \\
    &\quad \underbrace{5.34 \times 10^{58}}_{\text{EVs}} \times
       \underbrace{8.58 \times 10^{7}}_{\text{natures}} \times
       \underbrace{2.33 \times 10^{14}}_{\text{items}} \times
       \underbrace{4.70 \times 10^{7}}_{\text{Tera}} \notag \\
    &\approx \boxed{10^{215}}.
    \label{eq:gen9-total}
\end{align}
The moveset factor is an upper bound (using Mew's movepool for all species); all other factors are exact. The only approximation is the final $\log_{10}$ rounding.

\subsection{Generation 1 OU (RBY) Team Configuration Space}
\label{sec:gen1-derivation}

Gen 1 lacks abilities, held items, natures, and Terastallization, but uses distinct mechanics for stats.

\paragraph{Species \textbf{(E)}.}
\begin{equation}
    \binom{151}{6} = 14{,}888{,}600{,}755 \approx 1.49 \times 10^{10}.
\end{equation}

\paragraph{Movesets \textbf{(E)}.}
Generation 1 contains 164 standard battle moves.
Mew can learn all of them via TMs/HMs; using the full 164-move pool for all 6 Pokémon:
\begin{equation}
    \binom{164}{4}^6 = 29{,}051{,}001^6 \approx 6.01 \times 10^{44}.
\end{equation}

\paragraph{Determinant Values (DVs) \textbf{(E)}.}
Attack, Defense, Speed, and Special DVs are each in $\{0,\ldots,15\}$; HP DV is determined by the others.
In competitive Gen~1 play, Defense, Speed, and Special DVs are always set to 15 (their maximum), as there is no strategic reason not to maximize them. The sole exception is Attack DV, which is sometimes set to 0 on special attackers to minimize self-damage from Confusion. Thus Attack DV $\in \{0, 15\}$ (2 values) and all other DVs are fixed, giving 2 choices per Pok\'emon:
\begin{equation}
    2^{6} = 64.
\end{equation}

\paragraph{Stat Experience \textbf{(E)}.}
Each stat has Stat Experience in $\{0,\ldots,65535\}$ (unsigned 16-bit integer), but in competitive Gen~1 play all Stat Experience values are always maxed at 65535 via full EV training---there is no strategic reason to leave any stat below maximum. This factor therefore collapses to 1 and does not contribute to the team space.

\paragraph{Gen 1 team space.}
\begin{align}
    \mathcal{T}_{\text{Gen1}}
    &= \underbrace{\binom{151}{6}}_{\mathbf{E}} \times
       \underbrace{\binom{164}{4}^6}_{\mathbf{E}} \times
       \underbrace{2^{6}}_{\mathbf{E}}
    \approx 1.49 \times 10^{10}
       \times 6.01 \times 10^{44}
       \times 64
    \approx \boxed{10^{57}}.
    \label{eq:gen1-total}
\end{align}

\subsection{Battle State Spaces}
\label{sec:battle-state}

\paragraph{Common components.}

\emph{HP \textbf{(A)}.}
Current HP is an integer from 0 to maximum; we approximate with a representative $\text{max-HP} = 300$, giving 301 states (0--300) per Pok\'emon:
\begin{equation}
    301^{12} \approx 5.53 \times 10^{29}. \quad \mathbf{(A)}
\end{equation}
This is the only approximation in the derivation: actual max HP is determined by species, IVs, and EVs (already counted in the team space) and ranges from 1 (Shedinja) to $\sim$714 (Blissey). The representative value of 300 falls within the typical competitive range and does not materially affect the order-of-magnitude estimate.

\emph{Gen 9 field conditions \textbf{(E)}.}
\begin{align*}
    \text{Weather:} &\quad 1 + 4 \times 8 + 3 = 36 \text{ states} \\
                    &\qquad \text{(None; Sun/Rain/Sand/Snow with 1--8 turns; Harsh Sun/Heavy Rain/Strong Winds).} \\
    \text{Terrain:} &\quad 1 + 4 \times 8 = 33 \text{ states (None; Electric/Grassy/Misty/Psychic with 1--8 turns).}
\end{align*}

\emph{Side conditions \textbf{(E upper bound)}.}
Per-side conditions, each independent:
\begin{align*}
    \text{Hazards:} &\quad 2 \times 4 \times 3 \times 2 = 48 \text{ states} \\
                    &\qquad \text{(Stealth Rock on/off, Spikes 0--3, Toxic Spikes 0--2, Sticky Web on/off).} \\
    \text{Screens:} &\quad 9^3 = 729 \text{ states (Reflect/Light Screen/Aurora Veil each 0--8 turns).} \\
    \text{Tailwind:} &\quad 5 \text{ states (off or 1--4 turns).} \\
    \text{Safeguard:} &\quad 6 \text{ states (off or 1--5 turns).} \\
    \text{Mist:} &\quad 6 \text{ states (off or 1--5 turns).}
\end{align*}
Per-side total: $48 \times 729 \times 5 \times 6 \times 6 = 6{,}298{,}560$. Both sides: $6{,}298{,}560^2 \approx 10^{13.6}$.

\emph{Terastallization state \textbf{(E)}.}
Each player may Tera at most once per battle; state $=$ (whether Tera used) $\times$ (which of 6 Pokémon Tera'd, if applicable):
\begin{equation}
    (1 + 6)^2 = 49. \quad \mathbf{(E)}
\end{equation}

\emph{Pseudo-weather \textbf{(E)}.}
Turn-counted field-level volatile conditions:
\begin{align*}
    \text{Trick Room:} &\quad 6 \text{ states (off; 1--5 turns remaining).} \\
    \text{Gravity:} &\quad 6 \text{ states.} \\
    \text{Magic Room:} &\quad 6 \text{ states.} \\
    \text{Wonder Room:} &\quad 6 \text{ states.} \\
    \text{Fairy Lock:} &\quad 3 \text{ states (off; 1--2 turns remaining).}
\end{align*}
\begin{equation}
    \mathcal{P}_{\text{Gen9}} = 6^4 \times 3 = 3{,}888. \quad \mathbf{(E)}
\end{equation}

\emph{Slot conditions \textbf{(E)}.}
Conditions tied to a battlefield position (slot), not to the Pok\'emon occupying it.
In singles, there are 2 slots (one per side):
\begin{align*}
    \text{Wish:} &\quad 2 \text{ (pending or not).} \\
    \text{Future Sight/Doom Desire:} &\quad 3 \text{ (off; 1 or 2 turns remaining).} \\
    \text{Healing Wish:} &\quad 2 \text{ (pending or not).}
\end{align*}
Per slot: $2 \times 3 \times 2 = 12$. Two slots: $12^2 = 144$.

\emph{Per-active volatile statuses \textbf{(E upper bound)}.}
Each active Pok\'emon may simultaneously carry multiple volatile conditions.
We systematically enumerate all volatiles from the Pok\'emon Showdown source code (\texttt{conditions.ts}, \texttt{moves.ts}, \texttt{abilities.ts}), grouping mutually exclusive states.

\textbf{Independent binary/counter volatiles} (can all coexist).
We restrict to moves available in Generation~9; moves marked \texttt{isNonstandard:\ "Past"} in the Showdown source are excluded.\footnote{Excluded Past moves: Rage, Nightmare, Embargo, Heal Block, Octolock, Telekinesis, Foresight, Miracle Eye, Snatch, Grudge, Magic Coat, Laser Focus, Bide. These could technically be invoked via Metronome, but we adopt the same competitive-play framing used elsewhere in this derivation.}
Confusion (5: off or 2--5 turns),
Attract (2),
Leech Seed (2),
Substitute (2),
Taunt (5: off or 1--4 turns),
Encore (4: off or 1--3 turns),
Disable (5: off or 1--4 turns),
Torment (2),
Perish Song (4: off or 1--3 counter),
Aqua Ring (2),
Ingrain (2),
Focus Energy (2),
Yawn (3: off or 1--2 turns),
Curse (2),
No Retreat (2),
Tar Shot (2),
Salt Cure (2),
Syrup Bomb (4: off or 1--3 turns),
Imprison (2),
Charge (2),
Minimize (2),
Defense Curl (2),
Stockpile (4: 0--3 layers),
Glaive Rush (2),
Gastro Acid (2),
Power Trick (2),
Transform (2),
Trapped (2: Mean Look/Block/Shadow Tag),
Throat Chop (3: off or 1--2 turns),
Lock-On/Mind Reader (2).
\begin{equation}
    \mathcal{V}_{\text{indep}} \approx 6.04 \times 10^{11}. \quad \text{(product of all 30 factors above)}
\end{equation}

\textbf{Mutually exclusive groups} (at most one state per group):

\emph{Move locks} (a Pok\'emon can be locked into at most one multi-turn move):
None (1), locked move/Outrage etc.\ (1--3 turns = 3), two-turn move/Fly/Dig (1), must recharge/Hyper Beam (1), Uproar (1--3 turns = 3), Rollout (1--5 hits = 5), Ice Ball (1--5 hits = 5) $= 1+3+1+1+3+5+5 = 19$ states.

\emph{Protection moves} (at most one per turn):
None (1) + Protect, Baneful Bunker, Burning Bulwark, King's Shield, Obstruct, Silk Trap, Spiky Shield, Endure (8 variants) $= 9$ states.

\emph{Grounding/levitation} (mutually exclusive):
None (1), Smack Down (1), Magnet Rise (1--5 turns = 5) $= 7$ states.

\emph{Ability-specific volatiles} (a Pok\'emon has exactly one ability, so at most one applies):
None (1), Flash Fire active (1), Unburden active (1), Slow Start (1--5 turns = 5), Protosynthesis active (1), Quark Drive active (1) $= 10$ states.

\emph{Self-applied one-turn effects} (only one move per turn):
None (1), Roost (1), Destiny Bond (1) $= 3$ states.

\emph{Type changes}:
Type override (Soak, etc.): None (1) + 18 types + typeless from Burn Up (1) $= 20$ states.
Type addition: None (1) + Grass from Forest's Curse (1) + Ghost from Trick-or-Treat (1) $= 3$ states.

\emph{Choice lock} (from Choice Band/Scarf/Specs):
None (1) + locked into move 1/2/3/4 (4) $= 5$ states. Independent of multi-turn move locks above.

\emph{Other per-active factors}:
Stall counter (2: consecutive protect tracking), partial trapping (8: off or 1--7 turns), flinch (2), Powder (2), Electrify (2).

Combining all groups per active Pok\'emon:
\begin{align}
    \mathcal{V}_{\text{Gen9}} &= \mathcal{V}_{\text{indep}} \times 19 \times 9 \times 5 \times 2 \times 20 \times 3 \times 8 \times 2 \times 7 \times 10 \times 3 \times 2 \times 2 \notag \\
    &\approx 8.33 \times 10^{20}; \quad \mathcal{V}_{\text{Gen9}}^2 \approx 10^{41.8}. \quad \mathbf{(E\text{ upper bound})}
\end{align}

\emph{Item and ability state changes \textbf{(E upper bound)}.}
Items may be consumed, knocked off, or swapped (Trick/Switcheroo): 3 states per Pok\'emon.
Abilities may be suppressed or changed (Gastro Acid, Skill Swap, Mummy): 2 states per Pok\'emon.
\begin{equation}
    3^{12} \times 2^{12} \approx 10^{9.3}.
\end{equation}

\emph{PP.}
We omit move PP from the state space count. In competitive play, PP depletion is strategically relevant only in niche stalling scenarios; the vast majority of games end well before any move runs out of PP. Including binary PP tracking ($2^{48} \approx 10^{14.4}$) would increase all totals by $\sim$14 orders of magnitude but would not meaningfully reflect the strategic state of typical competitive battles.

\emph{Non-volatile status conditions \textbf{(E upper bound)}.}
Each Pok\'emon carries at most one non-volatile status, which persists through switches.
In Gen~9 the possible states per Pok\'emon are:
Healthy~(1), Burn~(1), Freeze~(1), Paralysis~(1), Poison~(1), Toxic~(1), Sleep with turn counter 1--3~(3):
\begin{equation}
    \mathcal{C}_{\text{Gen9}} = 9^{12} \approx 2.82 \times 10^{11}. \quad \mathbf{(E\text{ upper bound})}
\end{equation}

\paragraph{Gen 9 OU (singles).}

\emph{Active Pokémon \textbf{(E)}.}
One of 6 active per side: $6 \times 6 = 36$ arrangements.

\emph{Stat stages \textbf{(E)}.}
2 active Pokémon × 7 modifiable stats (Atk, Def, SpA, SpD, Spe, Acc, Eva), each $\in [-6,+6]$ (13 values):
\begin{equation}
    13^{14} \approx 3.94 \times 10^{15}.
\end{equation}

\begin{align}
    \text{Gen 9 OU battle state}
    &= \mathcal{T}_{\text{Gen9}}^2 \times 36 \times 301^{12} \times 13^{14} \times \mathcal{C}_{\text{Gen9}} \notag \\
    &\quad \times 36 \times 33 \times 6{,}298{,}560^2 \times 49 \times \mathcal{P}_{\text{Gen9}} \times 12^2 \notag \\
    &\quad \times \mathcal{V}_{\text{Gen9}}^{2} \times 3^{12} \times 2^{12} \notag \\
    &\approx 10^{430} \times 10^{1.6} \times 10^{29.7} \times 10^{15.6} \times 10^{11.5} \notag \\
    &\quad \times 10^{1.6} \times 10^{1.5} \times 10^{13.6} \times 10^{1.7} \times 10^{3.6} \times 10^{2.2} \notag \\
    &\quad \times 10^{41.8} \times 10^{5.7} \times 10^{3.6} \notag \\
    &\approx \boxed{10^{564}}.
    \label{eq:gen9-ou-battle}
\end{align}

\paragraph{Gen 9 VGC (doubles).}

VGC is a doubles format: each player has \emph{two} Pokémon active simultaneously, with position (left/right) mattering for targeting.
Each player also chooses which 4 of 6 Pokémon to bring during team preview (hidden from opponent).

\emph{Team preview \textbf{(E)}.}
$\binom{6}{4}^2 = 15^2 = 225$ joint selections.

\emph{Active positions \textbf{(E)}.}
Ordered pairs of active Pokémon from the 4 brought per side (left/right positions distinct): $P(4,2)^2 = (4 \times 3)^2 = 144$.

\emph{Stat stages \textbf{(E)}.}
4 active Pokémon × 7 modifiable stats:
\begin{equation}
    13^{28} \approx 1.55 \times 10^{31}.
\end{equation}

Since only 4 Pok\'emon per side participate in each VGC battle, HP is counted over 8 (not 12), non-volatile status over 8, and Terastallization over the 4 brought per side: $(1+4)^2 = 25$.

VGC adds several doubles-specific side conditions not present in singles:
Fire/Water/Grass Pledge field effects (5 states each, turn-counted),
Quick Guard (2), Wide Guard (2), Crafty Shield (2), Mat Block (2).
The per-side total becomes $6{,}298{,}560 \times 5^3 \times 2^4 = 12{,}597{,}120{,}000$, so both sides: $\sim 10^{20.2}$.

Per-active volatiles gain doubles-specific additions:
Helping Hand (2), redirection---Follow Me/Rage Powder/Spotlight (mutually exclusive, 4 states), Dragon Cheer (2), Ally Switch (2).
Slot conditions have 4 slots (2 per side) instead of 2: $12^4 = 20{,}736$.

\begin{align}
    \text{Gen 9 VGC battle state}
    &= \mathcal{T}_{\text{Gen9}}^2 \times 225 \times 144 \times 301^{8} \times 13^{28} \notag \\
    &\quad \times 9^{8} \times 36 \times 33 \times \text{SC}_{\text{VGC}}^2 \times 25 \times \mathcal{P}_{\text{Gen9}} \times 12^4 \notag \\
    &\quad \times \mathcal{V}_{\text{VGC}}^{4} \times 3^{8} \times 2^{8} \notag \\
    &\approx 10^{430} \times 10^{2.4} \times 10^{2.2} \times 10^{19.8} \times 10^{31.2} \notag \\
    &\quad \times 10^{7.6} \times 10^{1.6} \times 10^{1.5} \times 10^{20.2} \times 10^{1.4} \times 10^{3.6} \times 10^{4.3} \notag \\
    &\quad \times 10^{89.7} \times 10^{3.8} \times 10^{2.4} \notag \\
    &\approx \boxed{10^{622}}.
    \label{eq:gen9-vgc-battle}
\end{align}

\paragraph{Gen 1 OU (singles).}

Gen 1 has no weather, terrain, entry hazards, or screens.
Status conditions per Pokémon: Healthy + BRN + FRZ + PAR + PSN + bad PSN (turn counter 1--15) + SLP (turn counter 1--7) $= 1+1+1+1+1+15+7 = 27$ states\footnote{Toxic poison in Gen 1 deals $\lfloor N/16 \rfloor \times \text{max-HP}$ damage on turn $N$, so the counter 1--15 produces 15 distinct in-battle trajectories.}; for 12 Pokémon: $27^{12} \approx 10^{17.2}$ \textbf{(E)}.
Stat stages: 2 active Pokémon × 6 stats (Attack, Defense, Special, Speed, Evasion, Accuracy), each $\in [-6,+6]$: $13^{12} \approx 10^{13.4}$ \textbf{(E)}.

\emph{Per-active volatile statuses \textbf{(E upper bound)}.}
Volatile conditions that apply only to active Pok\'emon (cleared on switch).
In Gen~1, Reflect and Light Screen are per-Pok\'emon volatiles rather than field-wide screens.
Per active Pok\'emon:

\textbf{Independent:}
Confusion (5: none or 1--4 turns),
Leech Seed (2),
Substitute (2),
Focus Energy (2),
Disable (29: off, or one of 4 moves $\times$ 1--7 turns),
Reflect (2),
Light Screen (2),
Mist (2),
Minimize (2),
Transform (2),
Rage (2).

\textbf{Mutually exclusive move locks:}
None (1), Thrash lock (1--3 turns = 3), two-turn move/Fly/Dig (1), must recharge (1), Bide (1--2 turns = 2), partial trapping lock (1--4 turns = 4) $= 12$ states.

\textbf{Other per-active:}
Partial trapping (defender side, 5: off or 1--4 turns),
residual damage counter (16: 0--15, tracks toxic/trap damage),
flinch (2).

\begin{equation}
    \mathcal{V}_{\text{Gen1}} = (5 \times 2^7 \times 29 \times 2^2) \times 12 \times 5 \times 16 \times 2 = 142{,}540{,}800 \approx 10^{8.2}.
\end{equation}
\begin{equation}
    \mathcal{V}_{\text{Gen1}}^{2} \approx 10^{16.3}. \quad \mathbf{(E\text{ upper bound})}
\end{equation}

\begin{align}
    \text{Gen 1 OU battle state}
    &= \mathcal{T}_{\text{Gen1}}^2
       \times 36
       \times 301^{12}
       \times 13^{12}
       \times 27^{12}
       \times \mathcal{V}_{\text{Gen1}}^{2} \notag \\
    &\approx 10^{114}
       \times 10^{1.6}
       \times 10^{29.7}
       \times 10^{13.4}
       \times 10^{17.2}
       \times 10^{16.3}
    \approx \boxed{10^{192}}.
    \label{eq:gen1-battle}
\end{align}

\subsection{Summary}

\begin{table}[H]
\centering
\setlength{\tabcolsep}{8pt}
\caption{
State space complexity across Pokémon formats and classical games.
All Pokémon components are exact \textbf{(E)} except HP values (marked \textbf{A}); status conditions and volatile statuses are exact upper bounds.
Team space values are upper bounds.
Battle state includes item consumption/knockout state and ability suppression/swap state; PP is omitted as it is rarely strategic in competitive play (see text).
Volatile statuses are enumerated from the Pok\'emon Showdown source code with mutual exclusion groups (protect variants, move locks, grounding/levitation, ability-specific volatiles, self-applied one-turn effects) properly handled.
Gen 9 OU and VGC share the same team pool; VGC's larger battle state reflects 4 simultaneously active Pokémon (vs.\ 2 in singles), hidden team preview selection, and doubles-specific side conditions and volatiles.
}
\label{tab:state-space-comparison}
\resizebox{\textwidth}{!}{
\begin{tabular}{@{}cccc@{}}
\toprule
\textbf{Format} & \textbf{Team Space (1 player)} & \textbf{Battle State Space} & \textbf{Notes} \\
\midrule
Gen 1 OU (RBY)   & $\sim 10^{57}$ & $\sim 10^{192}$ & Singles; no items/abilities/natures; competitive DV constraints \\
Gen 9 OU         & $\sim 10^{215}$ & $\sim 10^{564}$ & Singles; Terastallization \\
Gen 9 VGC        & $\sim 10^{215}$ & $\sim 10^{622}$ & Doubles; bring 4 of 6; team preview \\
\midrule
Chess            & ---             & $\sim 10^{47}$  & --- \\
Go               & ---             & $\sim 10^{170}$ & --- \\
\bottomrule
\end{tabular}
}
\end{table}

\subsection{Team Space in Human Metagame}

\begin{figure}[h!]
    \centering
    \includegraphics[width=0.9\linewidth]{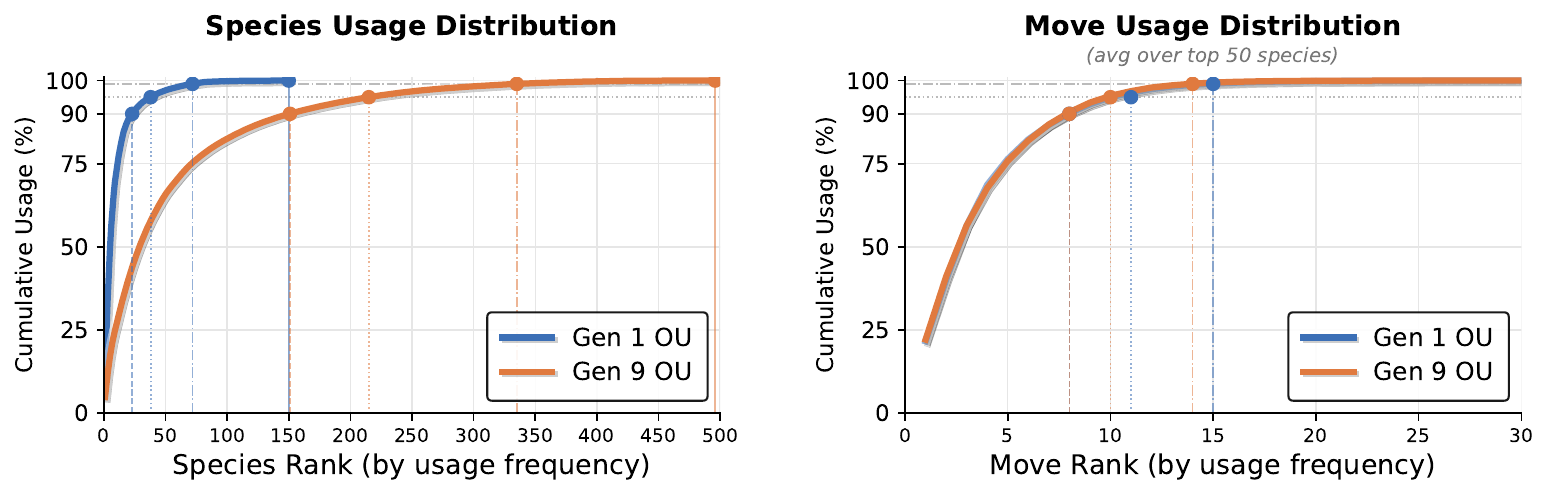}
    \caption{\textbf{Empirical usage distributions in Gen~1 OU and Gen~9 OU}. Derived from Smogon usage statistics (2014--2025,
rank~$\geq$1500).
\textbf{Left:} cumulative species usage---the fraction of all team-slot
appearances accounted for by the top-$k$ most-used species.
\textbf{Right:} cumulative move usage per species, averaged over the
top-50 species in each format.}
    \label{fig:team_complexity_usage_stats}
\end{figure}

The combinatorial space of Pokémon teams is vast when enumerating all \textit{legal} combinations of species, moves, items, abilities, and stat spreads (Table \ref{tab:state-space-comparison}). In practice, however, the effective space considered by competitive players is considerably smaller. Many options are dominated by clearly superior alternatives. For example, two moves may share the same type and role, but if one is less powerful and reliable, there is no competitive reason to justify its selection. Similar dominance relationships apply to items, abilities, and stat spreads. Table \ref{tab:gen1-vs-gen9-team-metrics} estimates this state-space reduction by restricting to options that Showdown players above a \textasciitilde median skill rating select in at least 1\% of battles. A second factor is metagame convergence. Even among viable options, competitive play concentrates around successful team archetypes as players adapt to one another. Figure \ref{fig:team_complexity_usage_stats} provides an example: a small fraction of available Pokémon account for the majority of human choices. Nevertheless, the number of competitively viable combinations is exceptionally large, and team design in competitive Pokémon remains a daunting optimization problem.

\begin{table}[h!]
\centering
\caption{\textbf{Effective team space on Showdown Gen~1 OU and Gen~9 OU ladders.}
Derived from public usage statistics (2014--2025, Glicko-1~$\geq$1500).
Only options appearing on at least 1\% of teams (for species) or 1\% of
that species' sets (for moves, items, abilities, and EV/nature spreads)
are counted.}
\label{tab:gen1-vs-gen9-team-metrics}
\resizebox{0.5\textwidth}{!}{
\begin{tabular}{@{}lcc@{}}
\toprule
\textbf{} & \textbf{Gen 1 OU} & \textbf{Gen 9 OU} \\
\midrule
Species in pool & 45 & 117 \\
\midrule
Avg.\ moves per species & 13.0 & 14.0 \\
Avg.\ items per species & --- & 6.6 \\
Avg.\ abilities per species & --- & 1.7 \\
Avg.\ EV/nature spreads per species & --- & 7.8 \\
\midrule
Team state space ($\log_{10}$) & 23.5 & 37.8 \\
\bottomrule
\end{tabular}
}
\end{table}

\section{Extended Discussion}
\label{sec:extended-discussion}

\subsection{Additional Findings}

\paragraph{Vision Remains a Fundamental Limitation.} Community observations consistently identified vision as the primary failure mode: ``Claude's vision is beyond fixing with a prompt---it barely knows what up and down are.'' This suggests that game-playing benchmarks will continue to challenge AI systems until vision-language integration improves substantially. Concurrent independent work comparing Gemini 3 Pro and 2.5 Pro on Pokémon Crystal~\citep{zhang2025gemini3playspokemon} corroborates this finding, reporting that vision-related errors remained a dominant failure mode across model generations despite improvements in reasoning capability.

\paragraph{Frontier vs.\ Open-Source Performance Gap.} Our baseline evaluation (Figure~\ref{fig:track1_baselines}) reveals that frontier models significantly outperform open-source alternatives in direct prompting, with Gemini 3 Flash achieving 71\% GXE compared to 29\% for Gemma3-1B without a harness. However, the PokeChamp harness substantially narrows this gap: Gemma3-1B with a harness reaches 53\% GXE, demonstrating that architectural support can partially compensate for raw model capability. This finding has practical implications for benchmark accessibility---researchers without frontier model access can still achieve meaningful results through careful harness design. The comparison also validates Pokémon as an OOD evaluation: if models had seen extensive Pokémon training data, we would expect the performance gap between model scales to be smaller, as smaller models would have relevant cached knowledge. Instead, the gap reflects genuine reasoning capability differences that a harness can only partially bridge.

\paragraph{Model Behavior Diversity.} Different model families exhibited distinct failure patterns:
\begin{itemize}
    \item Claude models showed ``memory corruption cascades''---once false information entered context, they followed incorrect paths for extended periods
    \item Gemini models exhibited ``roadblock'' behavior---oscillating between contradictory goals when facing conflicting objectives
    \item GPT models demonstrated excessive plan commitment---persisting with suboptimal strategies for hours/days
    \item Qwen models exhibited ``computational paralysis''---entering recursive damage calculation loops and getting stuck verifying basic type matchups (``water vs fire?'') while the battle state evolved. This failure mode is striking: \textit{in high-stakes sequential decision-making, extended deliberation can be catastrophic}
\end{itemize}

\begin{figure}[h]
    \centering
    \includegraphics[width=0.95\linewidth]{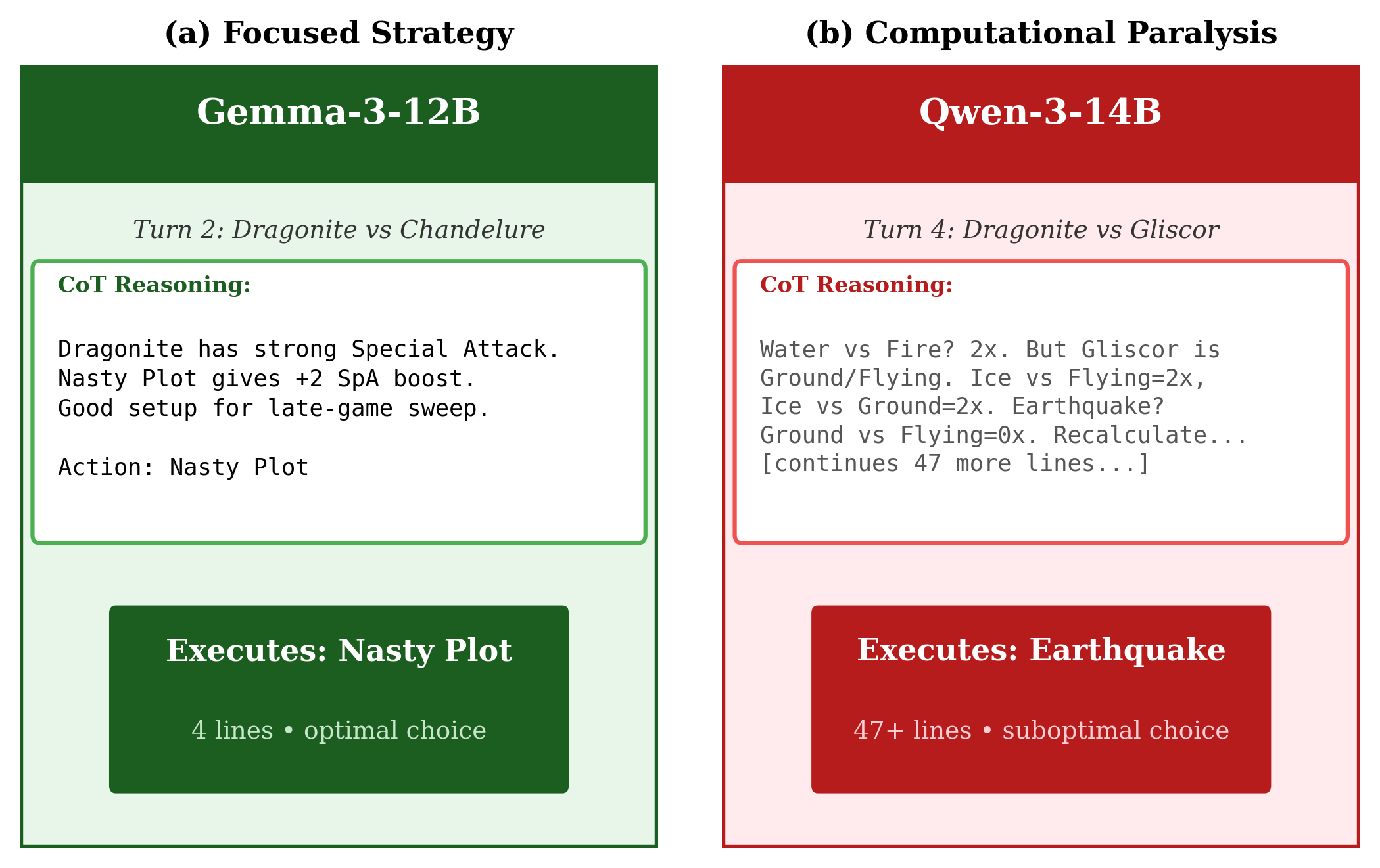}
    \caption{\textbf{Chain-of-Thought Visualization Reveals Failure Modes.} Screenshots from our live ladder stream showing synchronized CoT reasoning during a Gen 9 OU battle between Gemma-3-12B (a) and Qwen-3-14B (b). In Turn 2, Gemma articulates a brief strategic assessment (``boosting Special Attack will be beneficial'') and executes Nasty Plot. By Turn 4, Qwen's reasoning trace fills the panel with recursive damage calculations---enumerating type matchups, computing exact damage ranges, and deliberating between moves---exhibiting the ``computational paralysis'' described in the text. This failure mode is invisible to win/loss statistics alone.}
    \label{fig:cot_visualization}
\end{figure}

\paragraph{Chain-of-Thought Visualization Enables Failure Diagnosis.} We deployed a live ladder stream that visualizes LLM chain-of-thought synchronized with battle state. This revealed failure modes invisible to outcome-only metrics. Figure~\ref{fig:cot_visualization} shows Gemma-3-12B vs Qwen-3-14B on Gen 9 OU: while Gemma articulates strategy and executes decisively, Qwen's reasoning trace shows real-time ``panic''---recursive verification loops that consume the decision window. Without CoT visualization, Qwen's poor performance would appear as generic ``weak play'' rather than a specific, diagnosable pathology.

\subsection{Broader Impact}

\subsubsection{Beyond Pokémon: From Game Agents to Coding Agents}

An unexpected outcome of the X Plays Pokémon has been the transfer of harness techniques to other domains. The modular context engineering harness that emerged as essential for game-playing agents has influenced the design of autonomous coding agents. Systems like Claude Code now incorporate similar harness patterns: persistent memory across sessions, hierarchical planning for complex tasks, and structured perception of codebases. Recent develops have entirely overlapped with autonomous gaming agents as coding agents have been used in 24/7 loops with the ``Ralph Wiggum'' extension and ``OpenClaw''. This convergence suggests that game-playing benchmarks serve as verifiable benchmarks for developing autonomous agent capabilities that transfer to practical applications.

\section{Author Roles and Contributions}
\label{sec:authors}

\paragraph{Individual Contributors.}
Seth Karten$^{1}$, Jake Grigsby$^{2}$, Tersoo Upaa$^{1}$

\paragraph{PokeAgent Winning Teams (alphabetical).}
Junik Bae$^{6}$,
Seonghun Hong$^{14}$,
Hyunyoung Jeong$^{14}$,
Jaeyoon Jung$^{14}$,
Kun Kerdthaisong$^{15}$,
Gyungbo Kim$^{9}$,
Hyeokgi Kim$^{14}$,
Yujin Kim$^{9}$,
Eunju Kwon$^{9}$,
Dongyu Liu$^{7}$,
Patrick Mariglia$^{8}$,
Sangyeon Park$^{9}$,
Benedikt Schink$^{12}$,
Xianwei Shi$^{7}$,
Anthony Sistilli$^{11}$,
Joseph Twin$^{13}$,
Arian Urdu$^{12}$,
Matin Urdu$^{12}$,
Qiao Wang$^{10}$,
Ling Wu$^{10}$,
Wenli Zhang$^{7}$,
Kunsheng Zhou$^{7}$

\paragraph{Advisory Board.}
Stephanie Milani$^{3,4}$, Kiran Vodrahalli$^{5}$, Amy Zhang$^{2}$, Fei Fang$^{3}$, Yuke Zhu$^{2}$, Chi Jin$^{1}$

\paragraph{Affiliations.}
$^{1}$Princeton University \quad $^{2}$UT-Austin \quad $^{3}$Carnegie Mellon University \quad $^{4}$New York University \quad $^{5}$Google DeepMind \\
$^{6}$Team Heatz \quad $^{7}$Team PA-Agent \quad $^{8}$Team FoulPlay \quad $^{9}$Team 4thLesson \quad $^{10}$Team Q \\
$^{11}$Team Anthonys \quad $^{12}$Team Hamburg \quad $^{13}$Team Porygon2AI \quad $^{14}$Team Deepest \quad $^{15}$Team August

\paragraph{Correspondence.} Seth Karten and Jake Grigsby \quad \{\texttt{sethkarten@princeton.edu, grigsby@cs.utexas.edu}\}

\end{document}